\documentclass[runningheads]{llncs}
\usepackage{graphicx}

\usepackage{tikz}
\usepackage{bbm}
\usepackage{comment}
\usepackage{amsmath,amssymb,amsfonts} 
\usepackage{color}
\usepackage{booktabs}
\usepackage{multirow}
\usepackage{flushend}
\usepackage{hyperref}
\usepackage{comment}
\usepackage[nocompress]{cite}
\usepackage{pifont}
\newcommand{\cmark}{\text{\ding{51}}}
\newcommand{\xmark}{\text{\ding{55}}}
\usepackage[accsupp]{axessibility}  
\usepackage{hyperref}
\hypersetup{colorlinks=true,linkcolor=magenta,filecolor=magenta,urlcolor=magenta,}

\begin{document}
\pagestyle{headings}
\mainmatter
\def\ECCVSubNumber{1544}  

\title{Rethinking Learning Approaches for Long-Term Action Anticipation} 


\titlerunning{Rethinking Learning Approaches for Long-Term Action Anticipation}
%
\author{Megha Nawhal\inst{1}
\and
Akash Abdu Jyothi\inst{1}\index{Jyothi, Akash Abdu}
\and
Greg Mori\inst{1,2}
}
\authorrunning{M. Nawhal et al.}
%
\institute{Simon Fraser University, Burnaby, Canada 
\and
Borealis AI, Vancouver, Canada}

\newcommand{\anticipatr}{\textsc{Anticipatr}}
\maketitle

\begin{abstract}
Action anticipation involves predicting future actions having observed the initial portion of a video. Typically, the observed video is processed as a whole to obtain a video-level representation of the ongoing activity in the video, which is then used for future prediction. We introduce \anticipatr\ which performs long-term action anticipation leveraging segment-level representations learned using individual segments from different activities, in addition to a video-level representation. We propose a two-stage learning approach to train a novel transformer-based model that uses these two types of representations to directly predict a set of future action instances over any given anticipation duration. Results on Breakfast, 50Salads, Epic-Kitchens-55, and EGTEA Gaze+ datasets demonstrate the effectiveness of our approach.
\keywords{Action Anticipation; Transformer; Long-form videos}
\end{abstract}

\section{Introduction}
The ability to envision future events is a crucial component of human intelligence which helps in decision making during our interactions with the environment. We are naturally capable of anticipating future events when interacting with the environment in a wide variety of scenarios. Similarly, anticipation capabilities are essential to practical AI systems that operate in complex environments and interact with other agents or humans (\textit{e.g.}, wearable devices~\cite{soran2015generating}, human-robot interaction systems~\cite{koppula2015anticipating}, autonomous vehicles~\cite{ma2019trafficpredict,yu2020spatio}).


Existing anticipation methods have made considerable progress on the task of near-term action anticipation~\cite{vondrick2016anticipating,gao2017red,mahmud2017joint,furnari2020rolling,furnari2019would,damen2018scaling,girdhar2021anticipative,dang2021msr} that involves predicting the immediate next action that would occur over the course of a few seconds. While near-term anticipation is a valuable step towards the goal of future prediction in AI systems, going beyond short time-horizon prediction has applicability in a broader range of tasks 
that involve long-term interactions with the environment. 
The ability to anticipate actions over long time-horizons is imperative for applications such as efficient planning in robotic systems~\cite{chang2020procedure,gammulle2019forecasting} and intelligent augmented reality systems.

\begin{figure}[t]
    \centering
    \includegraphics[width=0.9\textwidth]{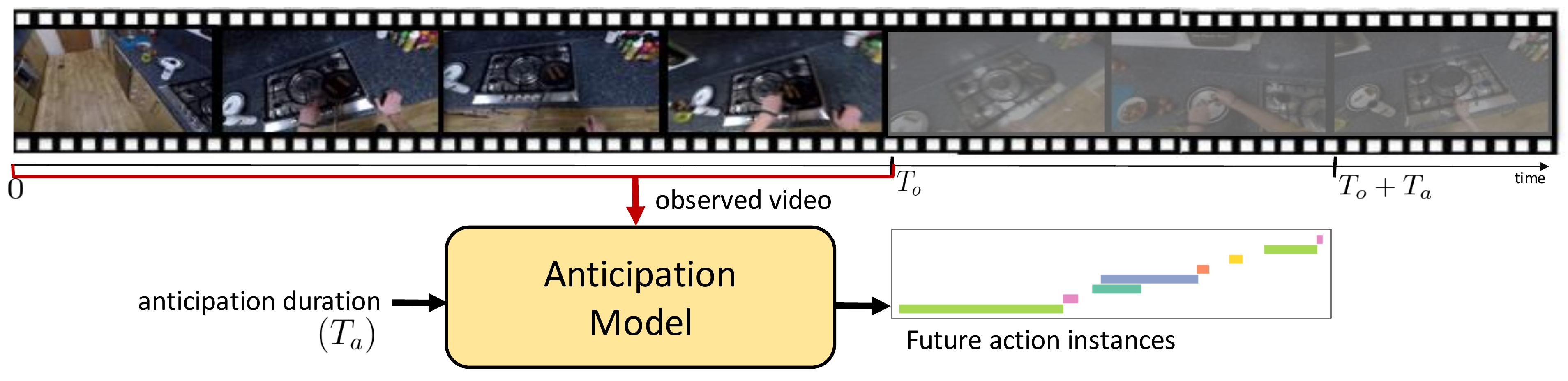}
    \caption{\textbf{Long-Term Action Anticipation.} Given the initial portion of an activity video ($0, \ldots, T_o$) and anticipation duration $T_a$, the task is to predict the actions that would occur from time $T_o+1$ to $T_o+T_a$. Our proposed anticipation model receives the observed video and the anticipation duration as inputs and directly predicts a set of future action instances. Here, the action anticipation is \textit{long-term} -- both the observed duration $T_o$ and the anticipation duration $T_a$ are in the order of minutes. 
    }
    \label{fig:teaser}
\end{figure}

In this paper, we focus on long-term action anticipation. Figure~\ref{fig:teaser} illustrates the problem -- having observed an initial portion of an untrimmed activity video, we predict \textit{what} actions would occur \textit{when} in the future. 

Long-term anticipation methods~\cite{farha2020long,abu2018will,gammulle2019forecasting,ke2019time,sener2020temporal} predict future actions based on the information in the observed video (\textit{i.e.}, an initial portion of an untrimmed activity video) that partially depicts the activity in the video. Current approaches rely on
encoding the observed video (input) as a whole to obtain \textit{video-level representations} to perform action anticipation. 

We propose a novel approach that leverages segment-level and video-level representations for the task of long-term action anticipation.
Consider the example in Figure~\ref{fig:teaser}. The video depicts the activity \textit{person making pasta} spanning several minutes. This activity has segments with actions such as \textit{slice onion}, \textit{put pesto}, \textit{put courgette}, \textit{add cheese}. One of these segments such as \textit{put pesto} tends to co-occur with actions involving objects such as \textit{courgette}, \textit{onion}, or \textit{cheese} in a specific order. 
However, other videos with a different activity, say, \textit{person making pizza}, could potentially have a similar set and/or sequence of actions in a different kitchen scenario. As such, while a specific sequence of actions (\textit{i.e.,} segments of a video) help denote an activity, an individual video segment (containing a  single action) alone contains valuable information for predicting the future.  
Based on this intuition, we introduce an approach that leverages segment-level representations in conjunction with video-level representations for the task of long-term action anticipation. In so doing, our approach enables reasoning beyond the limited context of the input video sequence.



In this work, we propose \anticipatr\ that consists of a two-stage learning approach employed to train a transformer-based model for long-term anticipation (see Fig.~\ref{fig:overview} for an overview). In the first stage, we train a \textit{segment encoder} to learn segment-level representations.
As we focus on action anticipation, we design this training task based on co-occurrences of actions. Specifically, we train the segment encoder to learn \textit{which future actions are likely to occur after a given segment}?
Intuitively, consider a video segment showing a pizza pan being moved towards a microwave. Irrespective of the ongoing activity in the video that contains this segment, it is easy to anticipate that certain actions such as \textit{open microwave}, \textit{put pizza} and \textit{close microwave} are more likely to follow than the actions \textit{wash spoon} or \textit{close tap}. 

In the second stage, we utilize both the segment-level and video-level representations for long-term action anticipation. We design a transformer-based model that contains two encoders: (1) the segment encoder to derive representations corresponding to segments in the observed video, and (2) a \textit{video encoder} to derive the video-level representations of the observed video. These encoded representations are then fed into an \textit{anticipation decoder} that predicts actions that would occur in the future. Our model is designed to directly predict a set of future action instances, wherein, each element of the set (\textit{i.e.}, an action instance) contains the start and end timestamps of the instance along with the action label. 
Using direct set prediction,
our approach predicts the actions at all the timestamps over a given anticipation duration in a single forward pass. 

To summarize, this paper makes the following contributions: (1) a novel learning approach for long-term action anticipation that leverages segment-level representations and video-level representations of the observed video, (2) a novel transformer-based model that receives a video and anticipation duration as inputs to predict future actions over the specified anticipation duration, (3) a direct set prediction formulation that enables single-pass prediction of actions, and (4) state-of-the-art performance on a diverse set of anticipation benchmarks: Breakfast~\cite{kuehne2014language}, 50Salads~\cite{stein2013combining}, Epic-Kitchens-55~\cite{damen2018scaling}, and EGTEA Gaze+~\cite{li2018eye}. 
Code is available at \href{https://github.com/Nmegha2601/anticipatr}{https://github.com/Nmegha2601/anticipatr}

Overall, our work highlights the benefits of learning representations that capture different aspects of a video, and particularly demonstrates the value of such representations for action anticipation. 




\section{Related Work}
\label{sec:relatedwork}

\noindent
\textbf{Action Anticipation.} Action anticipation is generally described as the prediction of actions before they occur. Prior research efforts have used various formulations of this problem depending on three variables: (1) anticipation format, \textit{i.e.}, representation format of predicted actions, (2) anticipation duration, \textit{i.e.}, duration over which actions are anticipated, and (3) model architectures.

Current approaches span a wide variety of anticipation formats involving different representations of prediction outcomes. They range from pixel-level representations such as frames or segmentations~\cite{liang2017dual,mathieu2015deep,luc2017predicting,bhattacharyya2018bayesian} and human trajectories~\cite{kitani2012activity,alahi2016social,jain2016structural,martinez2017human,hernandez2019human,dang2021msr} to label-level representations such as action labels~\cite{furnari2020rolling,farha2020long,lan2014hierarchical,gao2017red,vondrick2016anticipating,zeng2017visual,shi2018action,rodriguez2018action,furnari2019would,ke2019time,sener2020temporal,piergiovanni2020adversarial,rodin2022untrimmed,zhang2021weakly} or temporal occurrences of actions~\cite{abu2018will,gammulle2019forecasting,sun2019relational,liang2019peeking,mehrasa2019variational,mahmud2017joint} through to semantic representations such as affordances~\cite{koppula2015anticipating} and language descriptions of sub-activities~\cite{sener2019zero}. We focus on label-level anticipation format and use `action anticipation' to refer to this task.


Existing anticipation tasks can be grouped into two categories based on the anticipation duration: (1) near-term action anticipation, and (2) long-term action anticipation. In this paper, we focus on long-term action anticipation.

\textbf{Near-term anticipation} involves predicting label for the immediate next action that would occur in the range of a few seconds having observed a short video segment of duration of a few seconds. Prior work propose a variety of temporal modeling techniques to encode the observed segment such as regression networks~\cite{vondrick2016anticipating}, reinforced encoder-decoder network~\cite{gao2017red}, TCNs~\cite{zatsarynna2021multi}, temporal segment network~\cite{damen2018scaling}, LSTMs~\cite{furnari2019would,furnari2020rolling,osman2021slowfast},  VAEs~\cite{mehrasa2019variational,wu2021greedy} and transformers~\cite{girdhar2021anticipative}. 


\textbf{Long-term anticipation} involves predicting action labels over long time-horizons in the range of several minutes having observed an initial portion of a video (observed duration of a few minutes). A popular formulation of this task involves prediction of a sequence of action labels having observed an initial portion of the video. Prior approaches encode the observed video as a whole to obtain a video-level representation. Using these representations, these approaches either predict actions recursively over individual future time instants or use time as a conditional parameter to predict action label for the given single time instant. 
The recursive methods~\cite{abu2018will,farha2020long,gammulle2019forecasting,piergiovanni2020adversarial,sener2020temporal}
accumulate prediction error over time resulting in inaccurate anticipation outcomes for scenarios with long anticipation duration. The time-conditioned method~\cite{ke2019time} employs skip-connections based temporal models and aims to avoid error accumulation by directly predicting an action label for a specified future time instant in a single forward pass. However, this approach still requires multiple forward passes during inference as the task involves predicting actions at all future time instants over a given anticipation duration. Additionally, sparse skip connections used in \cite{ke2019time} do not fully utilize the relations among the actions at intermediate future time instants while predicting action at a given future time instant. In contrast to these approaches based on video-level representations, our approach leverages segment-level representations (learned using individual segments across different activities) in conjunction with video-level representations. Both these representations are utilized to directly predict action instances corresponding to actions at all the time instants over a given anticipation duration in a single forward pass. 

An alternate formulation of long-term anticipation proposed in ~\cite{nagarajan2020ego} focuses on predicting a set of future action labels without inferring when they would occur.~\cite{nagarajan2020ego} extracts a graph representation of the video based on frame-level visual affordances and uses graph convolutional network to encode the graph representation to predict a set of action labels. In contrast, our approach leverages both the segment-level and video-level representations of the input video and a transformer-based model to predict action instances - both action labels and their corresponding timestamps.

Other methods design approaches to model uncertainty in predicting actions over long time horizons~\cite{piergiovanni2020adversarial,abu2019uncertainty,ng2020forecasting} and self-supervised learning~\cite{qi2021self}. 
\noindent
\textbf{Early action detection.} The task of early action detection~\cite{ryoo2011human,hoai2014max,ma2016learning,shou2018online} involves recognizing an ongoing action in a video as early as possible given an initial portion of the video.
Though the early action detection task is different from action anticipation (anticipation involves prediction of actions \textit{before} they begin), the two tasks share the inspiration of future prediction.

\noindent
\textbf{Transformers in computer vision.} The transformer architecture~\cite{vaswani2017attention}, originally proposed for machine translation task, has achieved state-of-the-art performance for many NLP tasks. In recent years, there has been a flurry of work on transformer architectures designed for high-level reasoning tasks on images and videos. Examples include object detection~\cite{carion2020end}, image classification~\cite{dosovitskiy2021ViTR}, spatio-temporal localization in videos~\cite{girdhar2019video}, video instance segmentation~\cite{wang2020end}, action recognition~\cite{arnab2021vivit,zhang2021vidtr}, action detection~\cite{nawhal2021activity}, multi-object tracking~\cite{meinhardt2021trackformer}, next action anticipation~\cite{girdhar2021anticipative}, human-object interaction detection~\cite{zou2021end,kim2021hotr}. DETR~\cite{carion2020end} is a transformer model for object detection, wherein, the task is formulated as a set prediction problem. This work has since inspired transformer designs for similar vision tasks -- video instance segmentation~\cite{wang2020end} and human-object interaction detection~\cite{zou2021end}. Inspired by these works, we propose a novel  transformer architecture that uses two encoder to encode different representations derived from the input video and a decoder to predict the set of future action instances in a single pass. Our proposed decoder also receives anticipation duration as an input parameter to control the duration over which actions are predicted.

\section{Action Anticipation with \sc{Anticipatr}}
In this section, we first describe our formulation of long-term action anticipation and then describe our approach.


\noindent
\textbf{Problem Formulation. }
Let $\mathbf{v}_o$ 
be an observed video containing $T_o$ frames. Our goal is to predict the actions that occur from time $T_o+1$ to $T_o+T_a$ where $T_a$ is the anticipation duration, \textit{i.e.}, the duration over which actions are predicted. Specifically, we predict a set $\mathcal{A} = \{a^{i} = (c^{i},t_{s}^{i},t_{e}^{i}) \}$ containing future action instances. The $i$-th element denotes an action instance $a^{i}$ depicting action category $c^{i}$ occurring from time $t_{s}^{i}$ to $t_{e}^{i}$ where $T_o < t_s^{i} < t_e^{i} \leq T_o+T_a$. Here, $c^{i} \in \mathcal{C}$ where $\mathcal{C}$ is the set of action classes in the dataset.

Intuitively, for action anticipation, the observed video as a whole helps provide a broad, video-level representation of the ongoing activity depicted in the video. However, the observed video is composed of several segments that individually also contain valuable information about future actions and provide an opportunity to capture the video with segment-level representations. Using this intuition, in this paper, we propose \anticipatr\ that leverages these two types of representations of the observed video for the task of long-term anticipation.

\begin{figure}[t]
    \centering
    \includegraphics[width=0.66\textwidth]{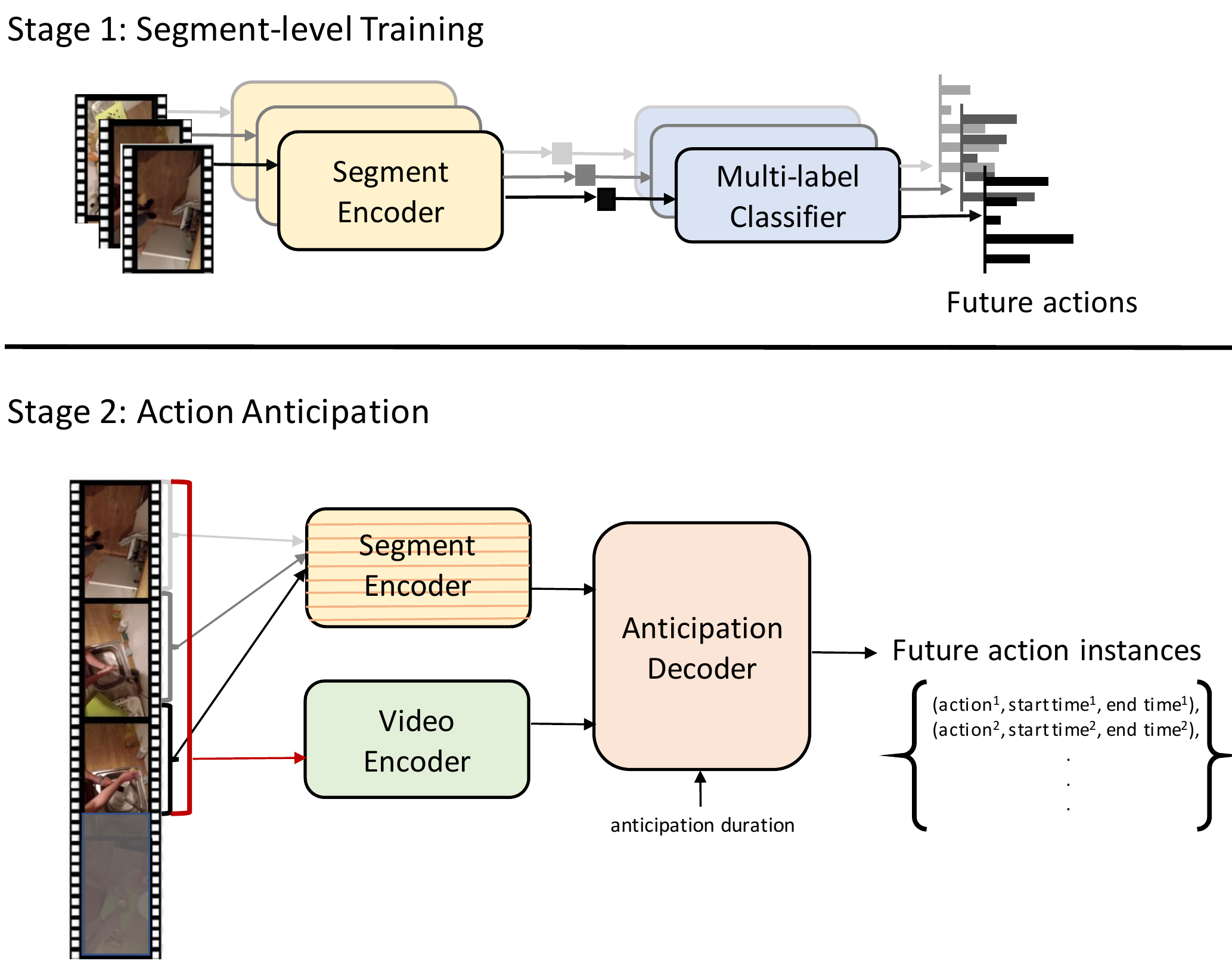}
    \caption{\textbf{Learning Approach.} \anticipatr\ uses a two-stage learning approach. In the first stage, we perform segment-level training (refer to Sec \ref{sec:pretraining}). Given a segment as input, we train a segment encoder to predict the set of action labels that would occur at any time after the occurrence of the segment in the activity video. In the second stage, we perform long-term action anticipation (refer to Sec \ref{sec:anticipation}). 
    We use video encoder to obtain video-level representation and segment encoder (trained in the first stage) is used to obtain segment-level representation. The anticipation decoder receives these two representations of the observed video to directly predict a set of action instances that would occur in the future over a given anticipation duration.
    }
    \label{fig:overview}
\end{figure}

\anticipatr\ employs a two-stage learning approach to train a transformer-based model that takes an observed video as input and produces a set of future action instances as output. See Fig.~\ref{fig:overview} for an overview. 
In the first stage, we train a \textit{segment encoder} that receives a segment (sequence of frames from a video) as input and predicts the set of action labels that would occur at any time in the future after the occurrence of the segment in the video. We refer to this stage as segment-level training (described in Sec. \ref{sec:pretraining}).  As the segment encoder only operates over individual segments, it is unaware of the broader context of the activity induced by a specific sequence of segments in the observed video. 

In the second stage, we train a \textit{video encoder} and an \textit{anticipation decoder} to be used along with the segment encoder for long-term action anticipation. The video encoder encodes the observed video to a video-level representation. The segment encoder (trained in the first stage) is fed with a sequence of segments from the observed video as input to obtain a segment-level representation of the video. The anticipation decoder receives the two representations along with the anticipation duration to predict a set of future action instances over the given anticipation duration in a single pass. The video encoder and anticipation decoder are trained using classification losses on the action labels and two temporal losses ($L_1$ loss and temporal IoU loss) on the timestamps
while the segment encoder is kept unchanged. We refer to this second stage of training as action anticipation (see Sec. \ref{sec:anticipation}).

\begin{figure}[t]
    \centering
    \includegraphics[width=0.9\textwidth]{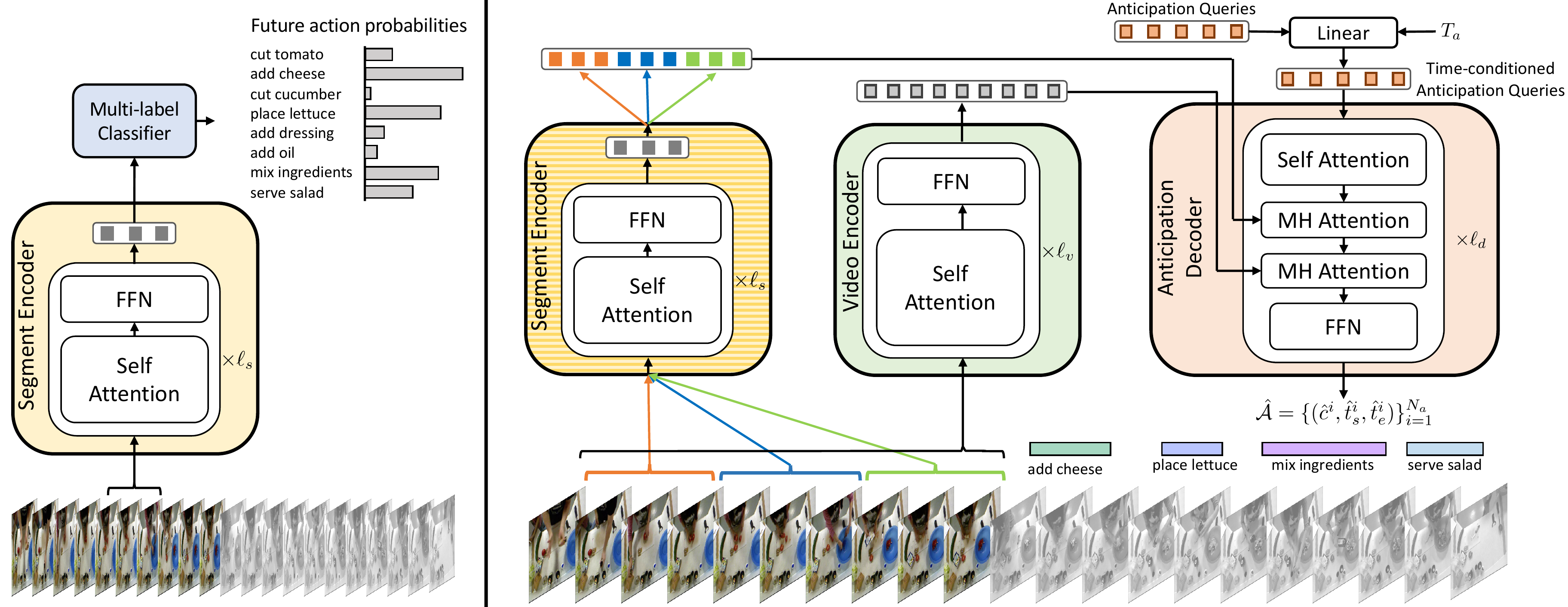}
    \caption{\textbf{Model Architecture.} Our model comprises three networks: \textit{segment encoder}, \textit{video encoder} and \textit{anticipation decoder} and is trained for long-term action anticipation in two stages. (\textit{left}) Segment-level training (Sec. \ref{sec:pretraining}): The segment encoder receives a segment as input and predicts a set of action labels that would occur at any time in the future (after the occurrence of segment in the video). (\textit{right}) Action Anticipation (Sec. \ref{sec:anticipation}): The video encoder encodes the observed video to a video-level representation. Concurrently, the video is divided into a sequence of segments and each segment is fed into the segment encoder (trained in first stage)
    The anticipation decoder receives the two representations along with an anticipation duration as inputs to directly predict a set of future action instances over the given anticipation duration. [MH Attention: Multi-head Attention, FFN: Feed Forward Network.]
    }
    \label{fig:model}
\end{figure}

\subsection{Stage 1: Segment-level Training}
\label{sec:pretraining}
In this stage, the segment encoder is trained on a segment-level prediction task to learn representations for individual segments. See Fig.~\ref{fig:model} (\textit{left}) for an overview.

\noindent
\textbf{Segment Encoder.} We design the segment encoder network $E_s$ as a sequence of $\ell_s$ transformer blocks containing a multi-head self-attention module followed by layernorm and a feed forward network~\cite{vaswani2017attention}.
This network is trained on the task of segment-level action anticipation.


\noindent
\textbf{Training.} 
During training, the segment encoder receives a segment (sequence of frames from a video) as input and predicts the set of action labels that would occur at any time in the future (starting from the temporal boundary, \textit{i.e.,} end of the segment until the end of that video) without inferring when they would occur. Depending on the segment, there could be multiple actions occurring between the end of segment and end of video. Thus, we formulate this training task as a multi-class multi-label classification. 

The training data for the segment encoder is derived from the training set in the original video dataset containing videos with action annotations. These input segments are obtained using the action boundaries provided in the training set. We do not require any additional annotations. 
Formally, given a video $\mathbf{v}$ containing $T$ frames, a segment  $\mathbf{v}_{s}^{(t',t'')}$, spanning time indices $t'$ to $t''$ where $0 \leq t' < t'' < T$, is taken as input. For this segment, the target is a binary vector $\mathbf{c}_{s}$ (dimension $|\mathcal{C}|$)
corresponding to the action labels that occur after the temporal boundary of the segment until the end of the video ($[\mathbf{v}^{t''+1},\ldots,\mathbf{v}^{T}]$).

The segment encoder $E_s$ receives the segment $\mathbf{v}_{s}^{(t',t'')}$ along with positional encodings $\mathbf{p}_{s}^{(t',t'')}$ (details in supplementary). The output of the encoder is an embedding $\mathbf{h} = [\mathbf{h}^{1},\ldots,\mathbf{h}^{t''-t'+1}]$ of dimension $(t''-t'+1) \times d_s$ where $d_s$ is the channel dimension. The output embeddings are then averaged along time dimension and fed into a linear layer $F$ followed by a sigmoid activation $\sigma$ to obtain future action probabilities $\mathbf{\hat c}_{s}$ of dimension $|\mathcal{C}|$, expressed as:
\begin{equation}
\begin{split}
    \mathbf{h} &= E_s\big(\mathbf{v}_{s}^{(t',t'')}, \mathbf{p}_{s}^{(t',t'')}\big)\\
    \mathbf{\hat{c}}_{s} &= \sigma\left(F\Bigg(\frac{1}{t''-t'+1}\sum_{i=1}^{t''-t'+1}\mathbf{h}^{i}\Bigg)\right).
\end{split}
\end{equation}
Here, $\mathbf{\hat{c}}_{s}$ is the output of a multi-label classifier where each element $c^{j}_{s}$ of $\mathbf{\hat c}_{s}$ denotes probability of corresponding action category $j \in \mathcal{C}$.
This network is trained using binary cross entropy loss between the prediction vector $\mathbf{\hat{c}}_{s}$ and target vector $\mathbf{c}_{s}$. 
Once trained, the linear layer $F$ is discarded and the segment encoder $E_s$ is used to obtain segment-level representations for the action anticipation stage. 

\subsection{Stage 2: Action Anticipation}
\label{sec:anticipation}
In the second stage of our approach, 
we use an encoder-decoder model that contains two encoders: (i) the segment encoder from the first stage, and (ii) a video encoder that encodes the observed video as a whole. The outputs of these two encoders along with an anticipation duration are fed into an anticipation decoder  which uses the representations from the two encoders to predict a set of future action instances over the given anticipation duration. See Fig.~\ref{fig:model} (\textit{right}).


\noindent
\textbf{Video Encoder.} The video encoder receives an observed video containing $T_o$ frames. We denote the input as $\mathbf{v}_{o}= [\mathbf{v}^{1},\ldots,\mathbf{v}^{T_o}]$. 
We design the encoder network $E_v$ as a sequence of  $\ell_v$ transformer blocks~\cite{vaswani2017attention} containing a multi-head self-attention module followed by layernorm and feed forward network. The encoder receives the features corresponding to the observed video $\mathbf{v}_{o}$ as input. As the self-attention module is permutation-invariant, we provide additional information about the sequence in the form of sinusoidal positional encodings~\cite{vaswani2017attention} $\mathbf{p}_{o} = [\mathbf{p}^{1},\ldots,\mathbf{p}^{T_o}]$ (see supplementary for additional explanation). Here, each element in the positional encoding sequence is added to the corresponding element in the video features and then fed into the encoder block. The encoder models temporal relationships in the observed video and transforms the input sequence to a contextual representation $\mathbf{h}_{v} = [\mathbf{h}_{v}^{1},\ldots,\mathbf{h}_{v}^{T_o}]$, expressed as:
\begin{equation}
    \mathbf{h}_{v} = E_v(\mathbf{v}_{o},  \mathbf{p}_{o}).
\end{equation}

\noindent
\textbf{Encoding Video Segments.} Concurrent to the video encoder, the input video is divided into a sequence of segments using temporal sliding windows.  
Specifically, a temporal window of size $k$ starting from frame index $i$ obtains a segment $[\mathbf{v}^{i},\ldots, \mathbf{v}^{i+k-1}]$, which is fed to the segment encoder to obtain the outputs $\mathbf{h}_{s}^{i},\ldots,\mathbf{h}_{s}^{i+k-1}$. The starting index $i$ slides across time with $i \in \{1,k+1,2k+1,\ldots,(T_o-k+1)\}$ generating the temporal windows, where the window size $k$ is a hyperparameter. The outputs of the segment encoder for all temporal windows are concatenated to obtain $\mathbf{h}_{s} = [\mathbf{h}_{s}^{1},\ldots,\mathbf{h}_{s}^{T_o}]$. 
During implementation, the representations can still be obtained in one forward pass of the segment encoder by stacking segments along the batch dimension of the input.
This segment-level representation of the video is complementary to the video-level representation that encodes the ongoing activity in the video.


\noindent
\textbf{Anticipation Decoder.} Given the video-level and the segment-level representations, the decoder aims to predict a set of future action instances over a given anticipation duration. The predicted set contains action instances of the form $\textrm{(label, start time, end time)}$. The anticipation decoder receives the following inputs: (i) \textit{anticipation queries} $\mathbf{q}_0$, (ii) anticipation duration $T_a$ over which actions are to be predicted, (iii) encoded representation $\mathbf{h}_{v}$ from video encoder $E_v$, and (iv) encoded representation $\mathbf{h}_{s}$ from segment encoder $E_s$.

The anticipation queries contain $N_a$ elements, \textit{i.e.}, $\mathbf{q}_0 = [\mathbf{q}_0^{1},\ldots,\mathbf{q}_0^{N_a}]$, wherein each query is a learnable positional encoding (more details in supplementary). We consider $N_a$ as a hyperparameter that is constant for a dataset 
and is sufficiently larger than the maximum number of action instances to be anticipated per video in the overall dataset. Each query $\mathbf{q}_0^{i}$ is then fed into a linear layer (weights shared for all values of $i$) along with the anticipation duration $T_a$ to obtain time-conditioned anticipation queries $\mathbf{q}_a^{i}$ for $i=1,\ldots,N_a$. This time conditioning enables the anticipation decoder to predict actions over any specified anticipation duration. 

The decoder network $D$ consists of $\ell_d$ blocks, wherein, each block contains a cascade of attention layers. The first attention layer is the multi-head self-attention block which models relations among the anticipation queries. The second attention layer is a multi-head encoder-decoder attention layer that maps the queries and the segment-level representations from the segment encoder. And, the third attention layer is another multi-head encoder-decoder attention layer that maps the output of previous layer to the video-level representation corresponding to the input. This third attention layer is followed by a feedforward network. The output of the decoder $\mathbf{y} = [\mathbf{y}^{1},\ldots,\mathbf{y}^{N_a}]$ serves as a latent representation of the action instances in the videos, expressed as:
\begin{equation}
    \mathbf{y} = D(\mathbf{q}_a,\mathbf{h}_v,\mathbf{h}_s)
\end{equation}

The decoder output is used to predict the set of action instances $\hat{\mathcal{A}}= \{ \hat a^{i} = ( \hat c^{i}, \hat t_s^{i}, \hat t_e^{i})\}_{i=1}^{N_a}$. Each element in decoder output $\mathbf{y}^{i}$ is fed into a linear layer followed by softmax 
to obtain prediction probabilities $\hat p^i(c)$ where $c=1,\ldots,|\mathcal{C}|+1$ and $\hat c^{i}$ is the class corresponding to maximum probability.
The number of queries $N_a$ is larger than the maximum number of action instances per video in the dataset. Thus, we introduce an additional class label $\varnothing$ indicating no action. 
$\mathbf{y}^i$ is also fed into another feedforward network with ReLU to obtain corresponding start timestamps $\hat {t}_{s}^{i}$ and end timestamps $\hat t_{e}^{i}$. 


\noindent
\textbf{Training.} To compute the loss, we first align the predictions with the groundtruth set of action instances. This alignment is necessary as there is no fixed prior correspondence between the predicted and the groundtruth set of action instances. Here, the predicted set for any video contains $N_a$ action instances, but the size of groundtruth set $ \mathcal{A}$ varies based on the video and is smaller than the predicted set. Thus, we first pad the groundtruth set to make it the same size as the predicted set by adding $N_a - |\mathcal{A}|$ elements with label $\varnothing$ indicating no action. 
Then, we use a pair-wise greedy correspondence algorithm to align the groundtruth and predicted sets. Starting with the groundtruth instance having the longest duration, we match each groundtruth instance with the unmatched predicted instance that has the maximum temporal overlap with the groundtruth instance. This results in a one-to-one mapping for loss computation (more details in supplementary).  

Consider the output of the set correspondence module as $\gamma$ denoting the permutation of the predicted set of instances, \textit{i.e.}, the groundtruth action instance $a^{i}$ is matched to predicted instance $\hat a^{\gamma(i)}$ for $i = 1,\ldots,N_a$. Given this alignment, we compute loss $\mathcal{L}$ over all the matched pairs as a weighted combination of cross-entropy loss for classification, and two temporal losses: $L1$ loss and IoU loss ($\mathcal{L}_{iou}$) for prediction of segment timestamps, defined as: 
\begin{equation}
\label{eq:loss}
\begin{split}
\mathcal{L} = \sum_{i=1}^{N_a} \Big[
& -\log (\hat p^{\gamma(i)}(c^{i}))
 + \mathbbm{1}_{\{c^{i} \neq \varnothing\}} \lambda_{L1} ||s^{i} - \hat s^{\gamma(i)}||_{1}\\
& + \mathbbm{1}_{\{c^{i} \neq \varnothing\}} \lambda_{iou} \mathcal{L}_{iou}(s^{i},\hat s^{\gamma(i)}) \Big],
\end{split}
\end{equation}
where $\lambda_{iou}, \lambda_{L1} \in \mathbb{R}^{+}$ are hyperparameters, $s^{i} = [t^{i}_s, t^{i}_e]$, $\hat s^{\gamma(i)} = [\hat t_s^{\gamma(i)}, \hat t_e^{\gamma(i)}]$ and $\hat p^{\gamma(i)}(c^{i})$ is the probability of the groundtruth class $c^{i}$ for prediction $\gamma(i)$. The video encoder and anticipation decoder are jointly trained to minimize this loss. We do not fine-tune the segment encoder in this stage. 

\noindent 
\textbf{Inference.} 
During inference, the video encoder takes the observed video as input and the segment encoder takes the chunked video (\textit{i.e.}, non-overlapping segments of fixed length) as input. The inputs to the decoder are: (i) anticipation queries
$\mathbf{q}_0 = 1,\ldots,N_a$ (a constant, regardless of input), (ii) anticipation duration $T_a$ (varies based on the input video and the anticipation requirement), (iii) output representation from the video encoder, and (iv) output representation from the segment encoder. The decoder predicts a set of action instances. 
Thus, our approach allows us to build a model that can anticipate actions over any future duration in a single pass by simply controlling the input $T_a$ to the decoder as shown by results in Table~\ref{tab:ltaa_bf_salads}.

In summary, \anticipatr\ uses a two-stage learning approach to train a transformer-based model (consisting of two encoders and one decoder) to predict a set of future action instances over any given anticipation duration. Our approach aims to perform action anticipation with segment-level
representations learned using individual video segments in conjunction with video-level representations learned by encoding input video as a whole. Our model anticipates actions at all time instants over a given anticipation duration in a single forward pass by directly predicting a set of future action instances.

\section{Experiments}

We conducted extensive experiments and analysis to demonstrate the effectiveness of our proposed approach. 

\noindent
\textbf{Datasets.} We evaluate on four established benchmarks for this task. These datasets of untrimmed videos vary in scale, diversity of labels and video duration. 

\textbf{Breakfast}~\cite{kuehne2014language} contains 1,712 videos each depicting one of 10 breakfast activities and annotated with action instances spanning 48 different action classes. On average, a video contains 6 action instances and has a duration of 2.3 minutes. For evaluation, we report the average across 4 splits from the original dataset. 

\textbf{50Salads}~\cite{stein2013combining} contains 50 videos, each showing a person preparing a salad. On average, there are 20 action instances per video spanning 17 action classes and duration is 6.4 minutes. Following the original dataset, we report the average across 5-fold cross-validation in our evaluation.

\textbf{EGTEA Gaze+ (EGTEA+)}~\cite{li2018eye} contains egocentric videos of 32 subjects following 7 recipes in a single kitchen. Each video depicts the preparation of a single dish. Each video is annotated with instances depicting interactions (\textit{e.g.}, open drawer), spanning 53 objects and 19 actions.

\textbf{EPIC-Kitchens-55 (EK-55)}~\cite{damen2018scaling} contains videos of daily kitchen activities. It is annotated for interactions spanning 352 objects and 125 actions. It is larger than the aforementioned datasets, and contains unscripted activities.

We represent the input videos by feature representations used in the benchmarks (see supplementary for details).
\setlength{\tabcolsep}{4pt}
\renewcommand{\arraystretch}{0.95}
\begin{table}[t]
\centering
\caption{\textbf{Results (Breakfast and 50Salads).} We report the mean over classes accuracy for different observation/anticipation durations. Higher values indicate better performance. Note that ``Sener \textit{et al.}~\cite{sener2020temporal} (features+labels)" use action labels from a segmentation algorithm as additional input. Baseline results are from respective papers.}
\scalebox{0.75}{
\begin{tabular}{ll@{\hskip 2mm}c@{\hskip 5mm}c@{\hskip 5mm}c@{\hskip 5mm}c@{\hskip 5mm}c@{\hskip 5mm}c@{\hskip 5mm}c@{\hskip 5mm}c}
\toprule
& Observation ($\beta_o$)\ $\rightarrow$ & \multicolumn{4}{c}{20\%} & \multicolumn{4}{c}{30\%} \\
\cmidrule(lr){3-6}\cmidrule(lr){7-10}
& Anticipation ($\beta_a$)\ $\rightarrow$ & 10\% & 20\% & 30\% & 50\% & 10\% & 20\% & 30\% & 50\%\\
\midrule 
\multirow{7}{*}{\rotatebox[origin=c]{90}{\textbf{Breakfast}}}&RNN~\cite{abu2018will} & 18.1 &17.2 &15.9 &15.8 &21.6 &20.0 &19.7 &19.2\\
&CNN~\cite{abu2018will}&17.9 &16.3 &15.3 &14.5 &22.4 &20.12 &19.7 &18.7\\
&RNN~\cite{abu2018will} + TCN & \phantom{0}5.9& \phantom{0}5.6 &\phantom{0}5.5 &\phantom{0}5.1 &\phantom{0}8.9 &\phantom{0}8.9 &\phantom{0}7.6 &\phantom{0}7.7\\
&CNN~\cite{abu2018will} + TCN &  \phantom{0}9.8 &\phantom{0}9.2 &\phantom{0}9.1 &\phantom{0}8.9 &17.6 &17.1 &16.1 &14.4\\
&Ke \textit{et al.}~\cite{ke2019time} &18.4 &17.2 &16.4 &15.8 &22.7 &20.4 &19.6 &19.7\\
&Farha \textit{et al.}~\cite{farha2020long}&25.9 &23.4 &22.4 &21.5 &29.7 &27.4 &25.6 &25.2\\
& Qi \textit{et al.}~\cite{qi2021self}&25.6&21.0&18.5&16.0&27.3&23.6&20.8&17.3\\
& Sener \textit{et al.}~\cite{sener2020temporal} (features) & 24.2& 21.1&20.0&18.1&30.4&26.3&23.8&21.2\\
& Sener \textit{et al.}~\cite{sener2020temporal} (features+labels) &\textbf{37.4}&31.8&30.1&27.1 &39.8&34.2&31.9&27.9\\
&\sc{Anticipatr}\  \textbf{(Ours)}& \textbf{37.4}&\textbf{32.0}&\textbf{30.3}&\textbf{28.6}&\textbf{39.9}& \textbf{35.7}&\textbf{32.1}&\textbf{29.4}\\

\midrule
\multirow{7}{*}{\rotatebox[origin=c]{90}{\textbf{50Salads}}} &RNN~\cite{abu2018will} &30.1&25.4&18.7&13.5&30.8&17.2&14.8&\phantom{0}9.8\\
&CNN~\cite{abu2018will}&21.2&19.0&15.9&\phantom{0}9.8&29.1&20.1&17.5&10.9\\
&RNN~\cite{abu2018will} + TCN & 32.3&25.5&19.1&14.1&26.1&17.7&16.3&12.9\\
&CNN~\cite{abu2018will} + TCN&16.0&14.7&12.1&\phantom{0}9.9&19.2&14.7&13.2&11.2\\
&Ke \textit{et al.}~\cite{ke2019time} &32.5&27.6&21.3&15.9&35.1&27.1&22.1&15.6\\
&Farha \textit{et al.}~\cite{farha2020long}&34.8&28.4&21.8&15.2&34.4&23.7&18.9&15.9\\
& Sener \textit{et al.}~\cite{sener2020temporal}(features)&
25.5&19.9&18.2&15.1&30.6&22.5&19.1&11.2\\
& Sener \textit{et al.}~\cite{sener2020temporal}(features+labels) &34.7 & 26.3 & 23.7 & 15.7 & 34.5 &26.1 &22.7 &17.1\\
& Qi \textit{et al.}~\cite{qi2021self}&37.9&28.8&21.3&11.1& 37.5&24.1&17.1&09.1\\
& Piergiovanni \textit{et al.} ~\cite{piergiovanni2020adversarial} & 40.4 & 33.7 & 25.4 & 20.9 & 40.7 & 40.1 & 26.4 &19.2\\
&\sc{Anticipatr}\  \textbf{(Ours)}&\textbf{41.1}&\textbf{35.0}&\textbf{27.6}&\textbf{27.3}&\textbf{42.8}&\textbf{42.3}&\textbf{28.5}&\textbf{23.6}\\
\bottomrule
\end{tabular}}
\label{tab:ltaa_bf_salads}
\end{table}

\setlength{\tabcolsep}{4pt}
\renewcommand{\arraystretch}{0.98}
\begin{table}[t]
\centering
\caption{\textbf{Results (EK-55 and EGTEA+).} We report mAP values for \textsc{all} classes, \textsc{frequent} classes ($> 100$ action instances) and \textsc{rare} class ($< 10$ action instances). Following ~\cite{nagarajan2020ego}, we report the mAP values averaged over different observation durations. Higher values implies better performance. Baseline results are from respective papers.}
\scalebox{0.75}{
\begin{tabular}{l@{\hskip 3mm}ccc@{\hskip 4mm}ccc}
\toprule
Method & \multicolumn{3}{c}{\textbf{EK-55}} & \multicolumn{3}{c}{\textbf{EGTEA+}}\\
\cmidrule(lr){2-4}\cmidrule(lr){5-7}
& \textsc{All} & \textsc{Freq} & \textsc{Rare} & \textsc{All} & \textsc{Freq} & \textsc{Rare}\\
\midrule
RNN & 32.6&52.3&23.3&70.4&76.6&54.3\\
I3D~\cite{carreira2017quo} & 32.7&53.3&23.0&72.1&79.3&53.3\\
ActionVLAD~\cite{girdhar2017actionvlad} &29.8&53.5&18.6&73.3&79.0&58.6\\
Timeception~\cite{hussein2019timeception}& 35.6&55.9&26.1&74.1&79.7&59.7\\
VideoGraph~\cite{hussein2019videograph}& 22.5&49.4&14.0&67.7&77.1&47.2\\
\textsc{EGO-TOPO}~\cite{nagarajan2020ego} & 38.0&56.9&\textbf{29.2}&73.5&80.7&54.7\\
\sc{Anticipatr}\textbf{(Ours)}& \textbf{39.1}&\textbf{58.1}&29.1& \textbf{76.8}&\textbf{83.3}&\textbf{55.1}\\
\bottomrule
\end{tabular}}
\label{tab:ltaa_epic_egtea}
\end{table} 

\noindent 
\textbf{Evaluation.} To measure the performance of our model, we adopt the evaluation protocol followed by state-of-the-art methods for these benchmark datasets. 

For Breakfast and 50Salads, we report the mean over classes accuracy averaged over all future timestamps in the specified anticipation duration, \textit{i.e.}, dense prediction evaluation as defined in ~\cite{abu2018will,ke2019time,farha2020long}. We use $\beta_o\%$ of a full video as observation duration and predict the actions corresponding to following $\beta_a\%$ of the remaining video. As per the benchmarks, we sweep the values of $\beta_o \in \{20,30\}$ and $\beta_a \in \{10,20,30,50\}$ denoting different observation and anticipation durations respectively. 
Note that a single trained model is used for predicting at all these values of $\beta_o$ and $\beta_a$ by just varying the anticipation duration input to the decoder.
Since the metric is computed over a dense anticipation timeline, we first convert our model predictions (set of action instances) into a timeline and then compute mean over classes accuracy (details in supplementary). 

For EK-55 and EGTEA+, we compute a multi-label classification metric (mAP) over the target action classes as defined in ~\cite{nagarajan2020ego}.
$\alpha_o\%$ of each untrimmed video is given as
input to predict all action classes in the future $(100-\alpha_o)\%$ of the video, \textit{i.e.}, until the end of the video. We sweep values of
$\alpha_o \in \{25, 50, 75\}$ representing different observation durations. Since the metric is computed only over the future action classes, we take the union of the class labels of predicted action instances to compute mAP.

\noindent
\textbf{Comparison with state-of-the-art.} Table~\ref{tab:ltaa_bf_salads} shows the results for Breakfast and 50Salads datasets in the \textit{`no groundtruth labels'} setting~\cite{ke2019time,sener2020temporal}. The results show that our approach outperforms existing methods by a considerable margin for different observation/anticipation durations. For these benchmarks, the most similar approach to ours is Sener \textit{et al.}~\cite{sener2020temporal} where they propose self-attention methods for temporal aggregation for long-term video modeling. 
In the setting similar to ours where they use only visual features as input, our approach outperforms ~\cite{sener2020temporal} with up to 13\% improvement. Moreover, when they also use action labels from a segmentation algorithm as input, our approach is still competitive despite not using such additional inputs. In addition, the benefit of our approach is more apparent when the anticipation duration is longer.

Table~\ref{tab:ltaa_epic_egtea} shows results on the long-term action anticipation benchmarks for EK-55 and EGTEA+ datasets, as defined by~\cite{nagarajan2020ego}. The results show that our model achieves competitive results with the state-of-the-art method~\cite{nagarajan2020ego}. While this benchmark only considers prediction of future action labels, our results demonstrate that the segment prediction in our model acts as a beneficial auxiliary task for label prediction.

\noindent
\textbf{Impact of Segment-level Training.}
Our two-stage learning approach separately learns video-level representations and segment-level representations. 
To analyze the impact of such two-stage training, we design following experiments.

\begin{figure}[t]
    \centering
    \includegraphics[width=0.9\textwidth]{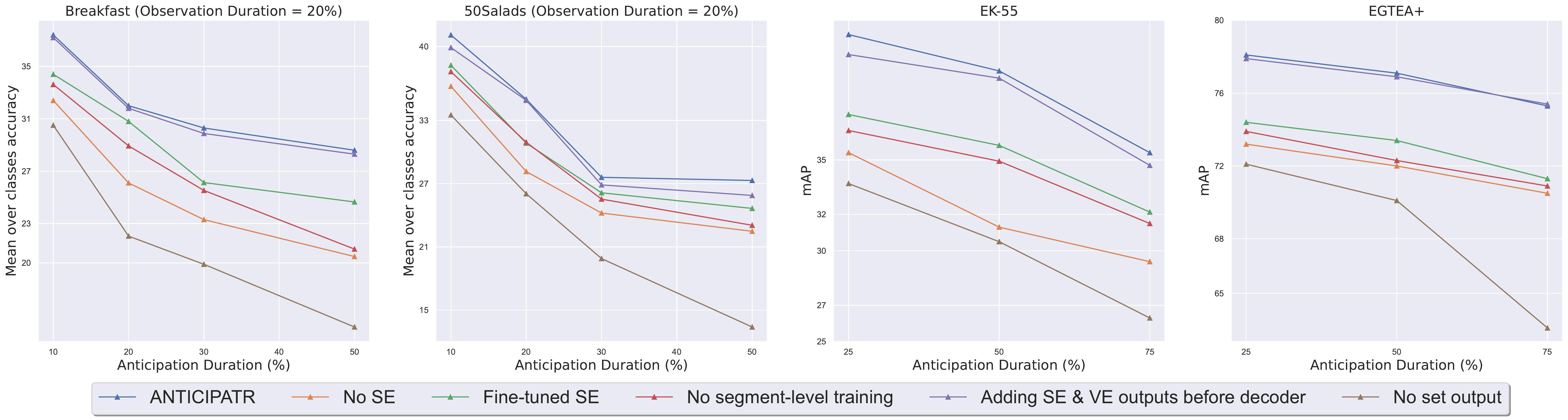}
    \caption{\textbf{Analysis.} Quantitative evaluation of the anticipation performance of ablated versions of \anticipatr. [SE: segment encoder; VE: video encoder].}
    \label{fig:ablation}
\end{figure}

(i) \textbf{Fine-tuned Segment Encoder.} In this experiment, we also fine-tune the segment encoder while training video encoder and decoder during the anticipation stage (Sec~\ref{sec:anticipation}). The results in Fig.~\ref{fig:ablation} (`Fine-tuned SE') indicate that fine-tuning the segment encoder hurts the anticipation performance. 
We believe fine-tuning the segment encoder with anticipation loss (Eq.~\ref{eq:loss}) perturbs the segment-level representation learned during first stage of training.

(ii) \textbf{No Segment-level Training.} In this experiment, we do not train the segment encoder network in a separate stage. Instead, we train all three networks (\textit{i.e.}, segment encoder, video encoder and anticipation decoder) jointly for the task of long-term action anticipation using the anticipation loss function (Eq.~\ref{eq:loss}). Here, the segment encoder receives videos chunked into short segments (same as the proposed two-stage training). However, it is directly tasked with solving a more difficult problem of simultaneously encoding segment-level representation and inferring its usage for long-term anticipation.
The results for all datasets presented in Fig.~\ref{fig:ablation} (`No Segment-level Training') illustrate that eliminating training of the segment encoder worsens the anticipation performance. This shows the value of learning the segment-level representations independently without being influenced by the overall activity in the input video.

In summary, these experiments demonstrate the importance of the two-stage learning approach and suggest that the two representations should be learned separately to serve their individual purposes during anticipation. 


\noindent
\textbf{Impact of Segment Encoder.} To evaluate the impact of learning segment-level representation, we conducted experiments without the segment encoder network. This ablated version only contains the video encoder and the anticipation decoder and is trained in a single-stage using the anticipation loss (Eq.~\ref{eq:loss}). The results in Fig.~\ref{fig:ablation} (`No SE') show that removing the segment-level representations considerably hurts the anticipation performance. This performance degradation is worse than just removing the segment-level training stage (`No segment-level training' in Fig.~\ref{fig:ablation}).
Thus, this experiment validates the benefit of the segment-level stream of information for action anticipation.

\noindent
\textbf{Impact of Set-based Output Representation. } 
In our approach, we model the anticipation output as a set of action instances.
We empirically validate this design by comparing with an alternative approach where the output is a sequence of action labels corresponding to the individual future time instants. We implement this by changing the anticipation queries (decoder input) during the anticipation stage -- we provide positional encodings corresponding to each time instant over anticipation duration and directly predict the labels corresponding to these time instants. While the prediction for all time instants still happens in a single pass, the decoder is required to transform a large number of anticipation queries. The results in Fig.~\ref{fig:ablation} (`No Set Output') show poor performance that worsen further as anticipation duration increases.
This is largely because the number of queries is too high for the decoder for effective modeling.

\noindent
\textbf{Fusion of Encoder Outputs.} To combine
the representation from segment encoder and video encoder, our model uses two encoder-decoder attention layers in the decoder blocks. We tested an alternative approach wherein we fused the representations using a simple addition along temporal dimension before feeding into the decoder. Here, we modify the decoder blocks to contain a single encoder-decoder attention layer. The results in Fig.~\ref{fig:ablation} (`Adding SE \& VE before decoder') indicate that this fusion approach leads to a slight decrease in anticipation performance. We believe adding the representations before decoder forces the computation of encoder-decoder attention weights by considering both information streams at once. In contrast, our \anticipatr\ approach of computing attention one-by-one enables it to first filter out the relevant information from segment-level representations learned across different activities and then contextualize them into the specific context of the input video.

\begin{figure}[t]
    \centering
    \begin{tabular}{cc}
    \includegraphics[width=0.48\textwidth]{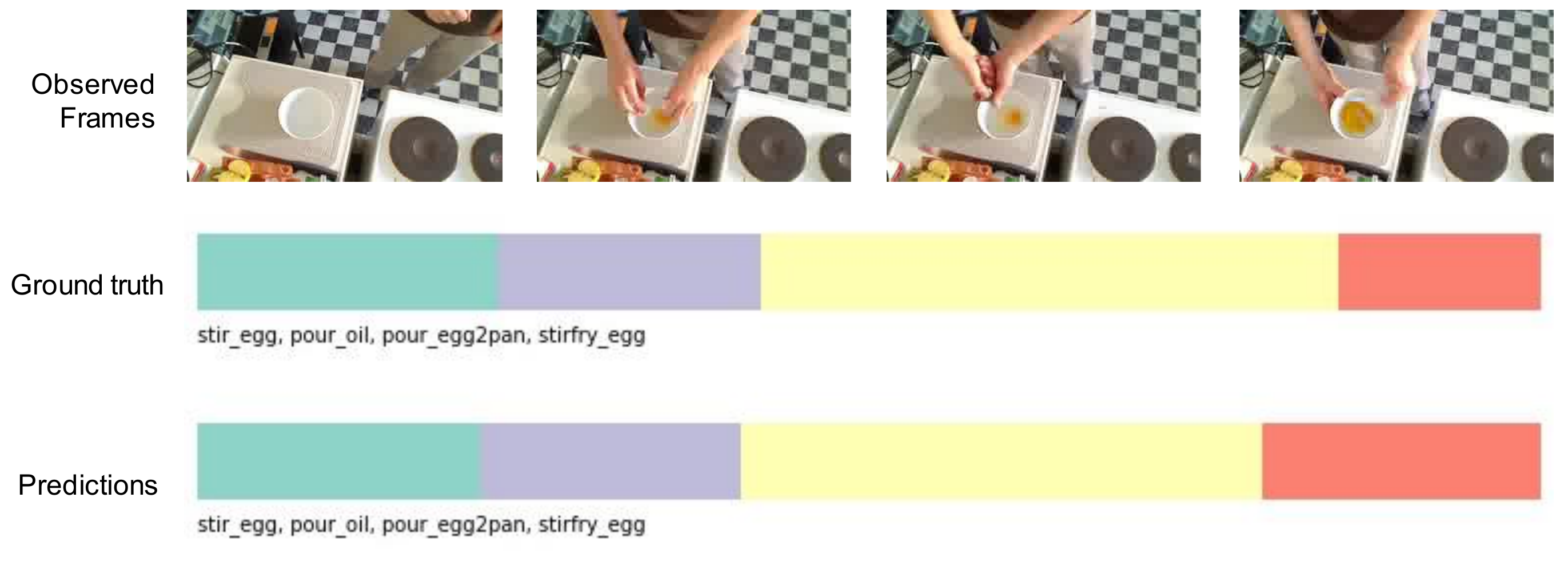}
    &
    \includegraphics[width=0.48\textwidth]{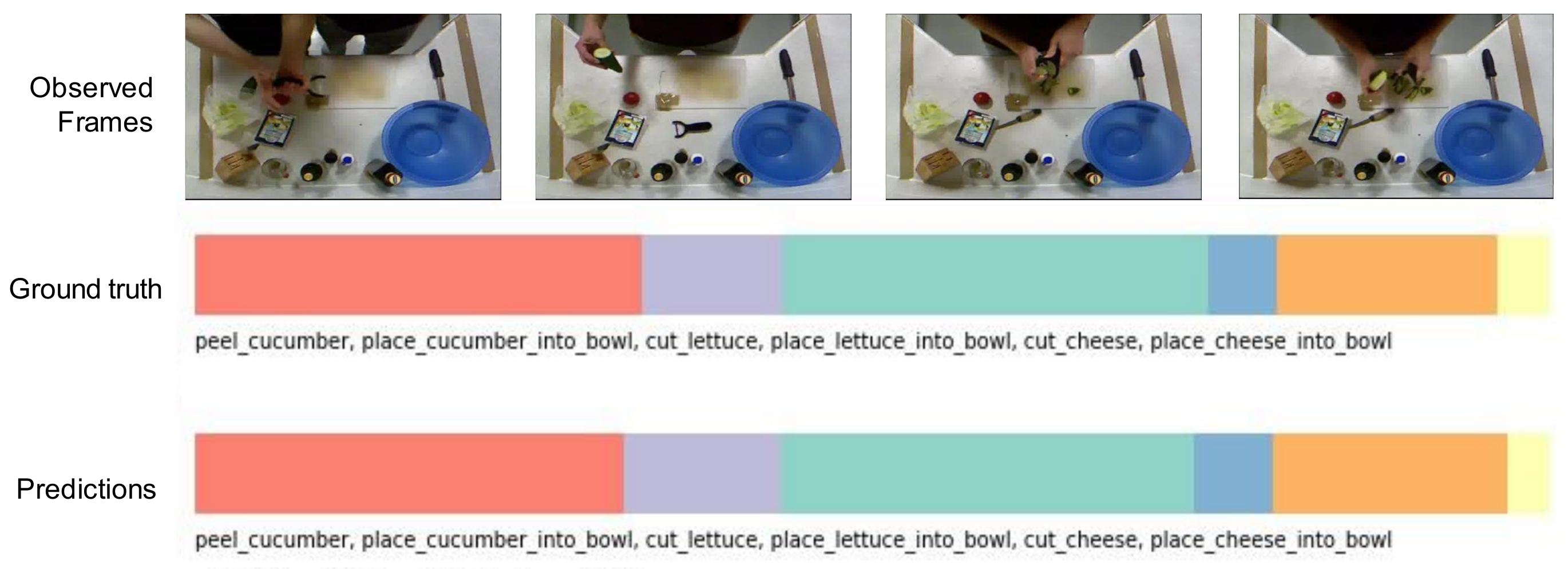}
    \end{tabular}
\caption{\textbf{Visualizations} from Breakfast (left) and 50salads (right) where 20\% of the video is observed and actions are anticipated over 50\% of the remaining video. 
}
\label{fig:qual_examples}
\end{figure}

\noindent
\textbf{Visualizations.} The examples in Fig.~\ref{fig:qual_examples} shows that our model effectively anticipates future actions. Please refer to supplementary material for additional visualizations and analysis of failure cases.

\section{Conclusion}
We introduced a novel approach for long-term action anticipation to leverage segment-level representations learned from individual segments across different activities in conjunction with a video-level representation that encodes the observed video as a whole. We proposed a novel two-stage learning approach to train a 
transformer-based model that receives a video and an anticipation duration as inputs and predicts a set of future action instances over the given anticipation duration. Results showed that our approach achieves state-of-the-art performance on long-term action anticipation benchmarks for Breakfast, 50Salads, Epic-Kitchens-55, and EGTEA Gaze+ datasets. Overall, our work highlights the benefits of learning representations that capture information across different activities for action anticipation.

\clearpage
\bibliographystyle{splncs04}
\bibliography{main}
\clearpage
\renewcommand\thesection{\Alph{section}}
\setcounter{section}{0}
\setcounter{table}{0}
\setcounter{secnumdepth}{3}
\setcounter{figure}{0}
\renewcommand{\thetable}{T\arabic{table}}
\renewcommand{\thefigure}{F\arabic{figure}}

\section{Appendix}
In this document, we provide additional quantitative and qualitative analyses, and additional details of the implementation of our approach. Specifically, this document contains the following items.

\begin{itemize}
    \item Sec.~\ref{sec:imp}: Technical details of the implementation and evaluation of our proposed approach
     \begin{itemize}
         \item Sec.~\ref{subsec:imparch}: Architecture details (network architectures and loss function)
         \item Sec.~\ref{subsec:imp_tech}: Implementation details (input representations and hyperparameters)
         \item Sec.~\ref{subsec:imp_eval}: Evaluation details 
    \end{itemize}
    \item Sec.~\ref{sec:ablation}: Additional ablation analysis
    \item Sec.~\ref{sec:qual}: Additional visualizations and qualitative analysis
    \item Sec.~\ref{sec:discussion} Additional discussion
\end{itemize}

\begin{figure*}
    \centering
    \begin{tabular}{c}
        \begin{tabular}{cc}
        \includegraphics[width=0.35\textwidth]{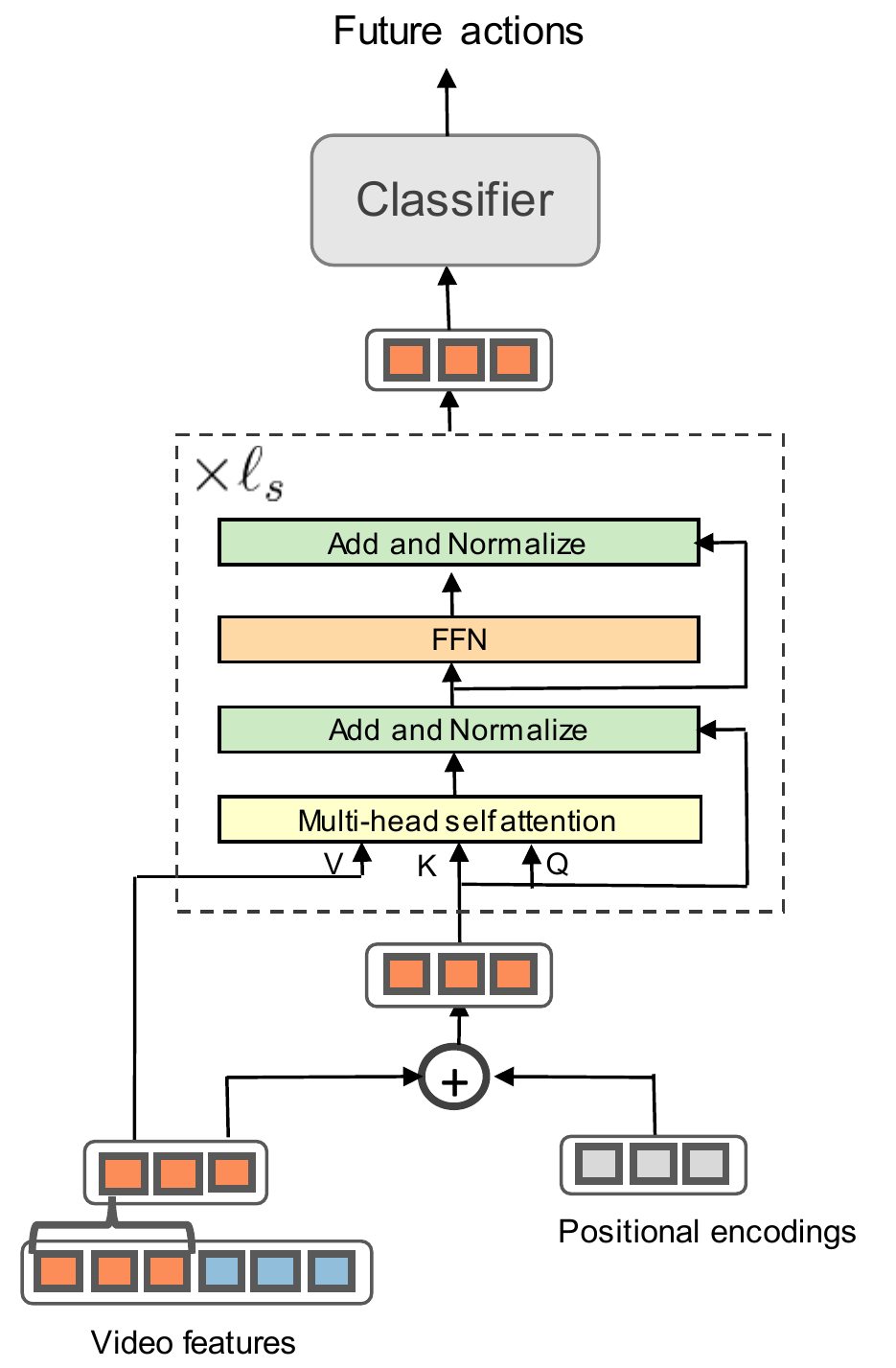}&
        \includegraphics[width=0.35\textwidth]{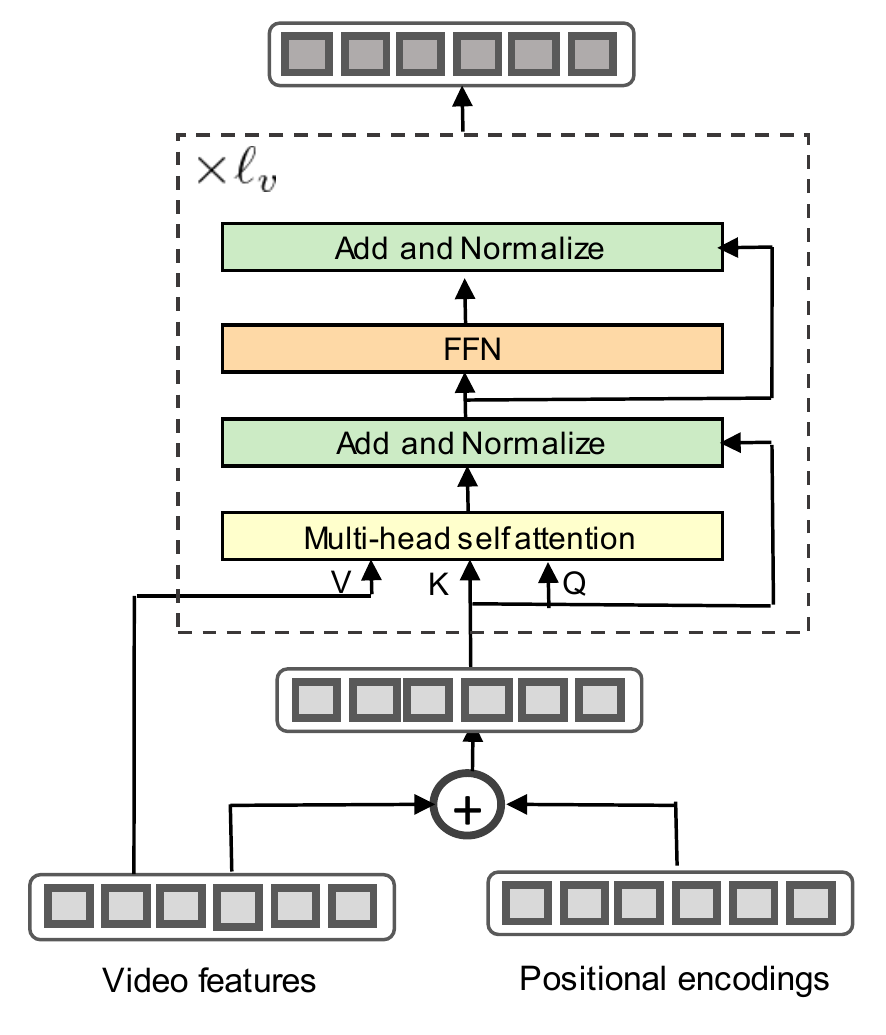}
        \\
        (a) & (b)\\
        \end{tabular}
        \\
        \\
    \includegraphics[width=0.75\textwidth]{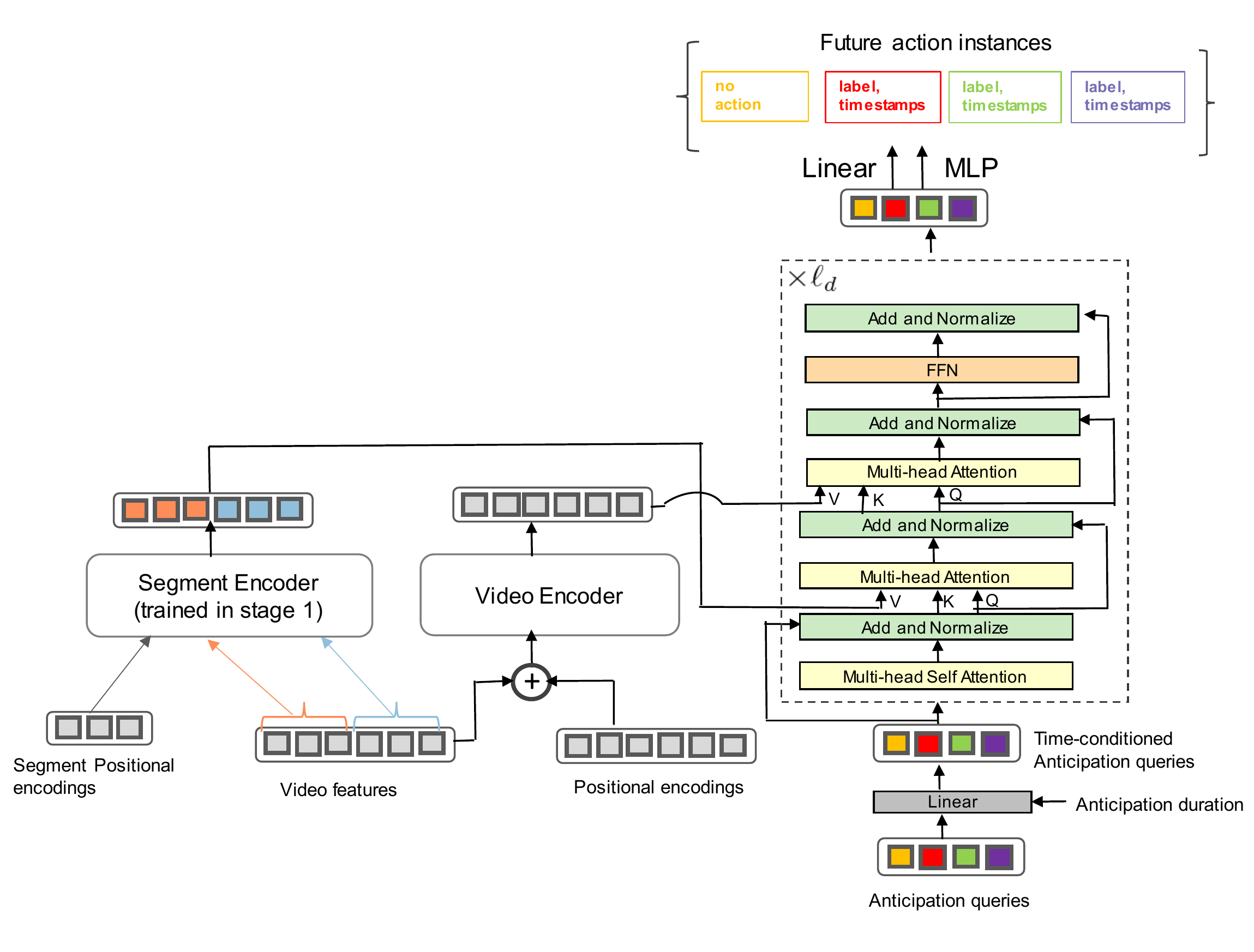}
    \\
    (c)\\
    \end{tabular}
    \vspace{0.3cm}
    \caption{\textbf{Detailed Architecture.} Architecture overview of (a) Segment encoder, (b) Video encoder, and (c) Anticipation Decoder. Refer to Sec.~\ref{subsec:imparch} for details. `Q',`K',`V' are query, key and value to the self-attention layer as described in ~\cite{vaswani2017attention}.}
    \label{fig:model_detailed}
\end{figure*}
\subsection{Technical Details}
\label{sec:imp}
In this section, we provide additional details for implementation of our proposed approach \anticipatr\ to supplement Sec. 3 in the main paper.

\subsubsection{Architecture Details}\hfill \break
\label{subsec:imparch}
We propose \anticipatr\ that uses a two-stage learning approach to train a transformer-based model for the task of long-term action anticipation. The model comprises three networks: \textit{segment encoder}, \textit{video encoder} and \textit{anticipation decoder}. 
Fig.~\ref{fig:model_detailed} shows the architecture of the three networks.

In the first stage, we train a \textit{segment encoder} that receives a segment (sequence of frames from a video) as input and predicts the set of action labels that would occur at any future time instant after the occurrence of the segment in the video. 

In the second stage, we train a \textit{video encoder} and an \textit{anticipation decoder} to be used along with the segment encoder for long-term action anticipation. The video encoder encodes the observed video to a video-level representation. The segment encoder (trained in the first stage) is fed with a sequence of segments from the observed video as input to obtain a segment-level representation of the video. The anticipation decoder receives the two representations along with the anticipation duration to predict a set of future action instances over the given anticipation duration in a single pass. The video encoder and anticipation decoder are trained using classification losses on the action labels and two temporal losses ($L_1$ loss and temporal IoU loss) on the timestamps while the segment encoder is kept unchanged.

    

\vspace{0.05in}
\noindent
\textbf{Positional Encoding for Segment Encoder.} The input to the segment encoder is a video segment. We represent the segment as a sequence of features. As the encoder is permutation-invariant, we provide temporal information in the segment using the sinusoidal positional encodings (c.f. Vaswani \textit{et al.}~\cite{vaswani2017attention}) based on timestamps corresponding to the features of input segment. Specifically, for each input feature of each embedding we independently use sine and cosine functions with different frequencies. We then concatenate them along the channel dimension to get the final positional encoding.
In our implementation, the embedding size is same as that of the segment feature so that they can be combined by simple addition of the positional encodings and segment features.

\vspace{0.05in}
\noindent
\textbf{Positional Encoding for Video Encoder.} The input to the video encoder is a video. We represent the video as a sequence of features. As the transformer encoder is permutation-invariant, we provide temporal information in the input video using the sinusoidal positional encodings (c.f. Vaswani \textit{et al.}~\cite{vaswani2017attention}) based on timestamps corresponding to the features of input video. Specifically, for each input feature of each embedding we independently use sine and cosine functions with different frequencies. We then concatenate them along the channel dimension to to get the final positional encoding.
In our implementation, the embedding size is same as that of the video feature so that they can be combined by simple addition of the positional encodings and video features.

\vspace{0.05in}
\noindent
\textbf{Anticipation Queries (Anticipation Decoder).} The anticipation queries are learnable positional encoding designed as a learnable embedding layer. The positional encoding layer receives integer index $i$ as input corresponding to $i-th$ anticipation query and provides an embedding $\mathbf{q}_0^{i}$ where $i \in \{1,\ldots,N_a\}$. In our implementation, we use \texttt{torch.nn.Embedding} in Pytorch to implement this. The weights of the layer are learnable during training, thus, the positional encoding layer is also learnable. The initialization of this layer requires maximum possible value of the index, \textit{i.e.}, $N_a$ in our case.

The anticipation queries $\mathbf{q}_0$ are then combined with anticipation duration $T_a$ using a simple neural network to create time-conditioned anticipation queries $\mathbf{q}_a$. These time-conditioned queries enable the model to predict actions over any specified anticipation duration.

\vspace{0.05in}
\noindent
\textbf{Training.} We provide supplemental details about computation of loss function used to train the networks in the second stage (\textit{i.e.}, action anticipation stage) of our \anticipatr\ approach. The training involves aligning groundtruth and predicted set of action instances followed by computing the anticipation loss over all aligned pairs.

\textbf{Greedy Set Correspondence.} Given an observed video, the groundtruth set of future action instances varies based on input whereas our anticipation decoder predicts a set of fixed size (larger than maximum size of groundtruth sets in the dataset). Therefore, there is no prior correspondence between the groundtruth and predicted set. We derive this correspondence using a greedy algorithm based on temporal overlap among instances. 
Intuitively, the objective is to correctly align actions at as many future time instants as possible. We first sort the action instances groundtruth set based on the descending order of the duration of the instances. We begin the alignment process with the groundtruth instance having the maximum duration. We lookup the predicted set to find the predicted instance that has maximum temporal overlap with this groundtruth instance. Since the predicted set is designed to represent a single action instance, the alignment between groundtruth and predicted set is one-to-one. Thus, to continue the alignment process, the matched groundtruth instance and predicted instance are removed from the corresponding sets. In this way, this process is repeated until the groundtruth set is empty. As the predicted set is of size larger than groundtruth set, the remaining predicted instances are mapped to $\varnothing$ denoting no action. In Sec.~\ref{sec:ablation}, we also evaluate anticipation results of models trained using another set correspondence algorithm, namely, Hungarian matcher (see Table~\ref{tab:ablation_matcher} and Table~\ref{tab:ablation_matcher_ek}).

\textbf{Loss function.} We compute loss $\mathcal{L}$ (defined in Eq. (4) in the main paper) over all the matched pairs as a weighted combination of cross-entropy loss for classification and two temporal losses ($L1$ loss and IoU loss $\mathcal{L}_{iou}$) for prediction of segment timestamps. 
Here, we provide our motivation behind temporal loss and provide additional description. 

The $L_1$ temporal loss is sensitive to the absolute value of the duration of the segments. The IoU loss $\mathcal{L}_{iou}$ is invariant to the duration of the segments. Thus, these two losses together are designed to incorporate different aspects of segment prediction. For completeness, we describe $\mathcal{L}_{iou}$ as follows.
\begin{equation}
     \mathcal{L}_{iou}(s^{i}, \hat s^{\gamma(i)}) = 1 - \frac{|s^{i} \cap \hat s^{\gamma(i)}|}{|s^{i} \cup \hat s^{\gamma(i)}|},
\end{equation}
 where $|.|$ is the duration of the instance, \textit{i.e.}, difference between end and start timestamp.

\subsubsection{Training Details}\hfill \break
\label{subsec:imp_tech}

For training of first stage, we use dropout probability of $0.1$. For the segment encoder, we use base model dimension as 2048 and set the number of encoder layers as 3 with 8 attention heads. We use an effective batch size of 64 for training segment encoder on this dataset. 
For training in the second stage, we use base model dimension in the video encoder and anticipation decoder as 2048 and set the number of encoder and decoder layers as 3 with 8 heads. 

We use four datasets -- Breakfast, 50Salads, EPIC-Kitchens-55, EGTEA Gaze+ -- to evaluate our model on long-term action anticipation. We provide dataset-specific hyperparameters as follows.

We train all our models using AdamW~\cite{loshchilov2017decoupled} optimizer on 4 Nvidia V100 32GB GPUs. We initialize all the learnable weights using Xavier initialization.

\textbf{Breakfast.} We represent input videos as I3D features provided by \cite{mstcn}. We choose $N_a$ (anticipation queries) to be 150. We use an effective batch size of 16 for training the video encoder and anticipation decoder on this dataset on the long-term anticipation task. We train our models with a learning rate of 1e-4 and a weight decay of 0. The model is trained for 4000k steps. We use a dropout probability of 0.1. We set $\lambda_{L1} = 3$ and $\lambda_{iou} = 5$. To obtain segment-level representation of the observed video during action anticipation, we use a temporal window of length $k = 16$.

\textbf{50Salads.} We represent input videos as Fisher vectors computed using~\cite{fv}. We choose $N_a$ (anticipation queries) to be 80. We use an effective batch size of 16 for training the video encoder and anticipation decoder on this dataset on the long-term anticipation task. We use a learning rate of 1e-5 and a weight decay of 1e-5. We train the model for 3000k steps and reduce the learning rate by factor of 10 after 1500k steps. We don't use dropout for this dataset. We set $\lambda_{L1} = 3$ and $\lambda_{iou} = 5$. To obtain segment-level representation of the observed video during action anticipation, we use a temporal window of length $k = 48$.

\textbf{EPIC-Kitchens-55.}
We represent input videos as I3D features provided by ~\cite{nagarajan2020ego,egotopo}. We use an effective batch size of 16 for training the video encoder and anticipation decoder in the second stage. We choose $N_a$ (anticipation queries) to be 900. We use a learning rate of 1e-4 and a weight decay of 1e-5. We train the model for 6000k steps and reduce the learning rate by factor of 10 after 4000k steps. We use a dropout probability of 0.1. We set $\lambda_{L1} = 5$ and $\lambda_{iou} = 8$. To obtain segment-level representation of the observed video during action anticipation, we use a temporal window of length $k = 32$.

\textbf{EGTEA Gaze+.}
We represent input videos as I3D features provided by ~\cite{nagarajan2020ego,egotopo}. We use an effective batch size of 16 for training the video encoder and anticipation decoder in the second stage. We choose $N_a$ to be 600. We use a learning rate of 1e-5 and a weight decay of 1e-5. We train the model for 4000k steps and reduce the learning rate by factor of 10 after 3000k steps. We use a dropout probability of 0.1. We set $\lambda_{L1} = 3$ and $\lambda_{iou} = 5$. To obtain segment-level representation of the observed video during action anticipation, we use a temporal window of length $k = 24$.

\subsubsection{Evaluation Details}\hfill \break
\label{subsec:imp_eval}
Note that our model predicts a set of action instances, wherein, each action instance is of the form $\text{(label, start time, end time)}$. To evaluate the model outputs as per the benchmarks, we do the following postprocessing.

For Breakfast and 50Salads, following the benchmark~\cite{sener2020temporal}, we evaluate the action anticipation outputs over a dense timeline. Our proposed \anticipatr\ predicts a set of action instances. During evaluation, we process this set of action instances to construct a timeline corresponding to the anticipation duration. We refer to the timeline as a sequence of action labels for time instants in the anticipation duration, \textit{i.e.}, between $T_o+1,\ldots,T_o+T_a$. In the benchmarks, the timeline contains a single action class corresponding to each time instant. We iterate over the predicted set to assign class labels to this timeline. Specifically, for each action instance in the predicted set, we assign the predicted action class to the time instants that are within the predicted segment (determined by predicted start and end timestamp). When predicted action instances overlap at certain time instants, we assign the action class with highest probability score among the overlapping predictions. Once the timeline is constructed, we compute mean over classes accuracy~\cite{sener2020temporal} to evaluate the model performance. Note that we are constructing this timeline only during evaluation to follow the benchmark evaluation protocols.

For EPIC-Kitchens-55 and EGTEA Gaze+, we perform a union over the action classes in the predicted set of instances to obtain a set of future action classes. We remove $\varnothing$ class from this set and use this set to compute mAP as described in benchmark \cite{nagarajan2020ego}.

\subsection{Additional Ablation Analysis}
\label{sec:ablation}
In this section, we report our findings from additional ablation experiments.
\setlength{\tabcolsep}{4pt}
\renewcommand{\arraystretch}{0.95}
\begin{table}[t]
\centering
\caption{\textbf{Ablation: Loss function (Breakfast and 50Salads).} We report the mean over classes accuracy for different observation/anticipation durations. Higher values indicate better performance. \cmark\ and \xmark\ indicate whether the component of the temporal loss is used or not respectively.}
\scalebox{0.54}{
\begin{tabular}{lll@{\hskip 5mm}c@{\hskip 5mm}c@{\hskip 5mm}c@{\hskip 5mm}c@{\hskip 5mm}c@{\hskip 5mm}c@{\hskip 5mm}c@{\hskip 5mm}c}
\toprule
& Method & $\beta_o$\ $\rightarrow$ & \multicolumn{4}{c}{20\%} & \multicolumn{4}{c}{30\%} \\
\cmidrule(lr){4-7}\cmidrule(lr){8-11}
& & $\beta_a$\ $\rightarrow$ & 10\% & 20\% & 30\% & 50\% & 10\% & 20\% & 30\% & 50\%\\
\midrule 
\textbf{Breakfast}
&$L_1$: \xmark; $\mathcal{L}_{iou}$:\cmark&&36.2&30.7&28.6&26.4&38.7&33.9&31.0&27.3\\
&$L_1$: \cmark; $\mathcal{L}_{iou}$:\xmark&&36.5&31.1&29.1&28.2&39.2&34.2&31.7&28.1\\
&$L_1$: \cmark; $\mathcal{L}_{iou}$:\cmark&& \textbf{37.4}&\textbf{32.0}&\textbf{30.3}&\textbf{28.6}&\textbf{39.9}& \textbf{35.7}&\textbf{32.1}&\textbf{29.4}\\
\midrule
\textbf{50Salads}
&$L_1$: \xmark; $\mathcal{L}_{iou}$:\cmark&&40.2&33.9&26.8&26.0&41.9&41.4&27.6&23.3\\
&$L_1$: \cmark; $\mathcal{L}_{iou}$:\xmark&&40.8&34.5&27.1&26.8&42.1&41.6&27.9&23.4\\
&$L_1$: \cmark; $\mathcal{L}_{iou}$:\cmark&&\textbf{41.1}&\textbf{35.0}&\textbf{27.6}&\textbf{27.3}&\textbf{42.8}&\textbf{42.3}&\textbf{28.5}&\textbf{23.6}\\
\bottomrule
\end{tabular}}
\label{tab:ablation_loss}
\end{table}

\setlength{\tabcolsep}{4pt}
\renewcommand{\arraystretch}{0.98}
\begin{table}[t]
\centering
\caption{\textbf{Ablation: Loss function (EK-55 and EGTEA+).} We report mAP values for \textsc{all} classes, \textsc{frequent} classes ($> 100$ action instances) and \textsc{rare} class ($< 10$ action instances). Following ~\cite{nagarajan2020ego}, we report the mAP values averaged over different observation durations. Higher values implies better performance. \cmark\ and \xmark\ indicate whether the component of the temporal loss is used or not respectively. }
\scalebox{0.7}{
\begin{tabular}{l@{\hskip 3mm}ccc@{\hskip 4mm}ccc}
\toprule
Method & \multicolumn{3}{c}{\textbf{EK-55}} & \multicolumn{3}{c}{\textbf{EGTEA+}}\\
\cmidrule(lr){2-4}\cmidrule(lr){5-7}
& \textsc{All} & \textsc{Freq} & \textsc{Rare} & \textsc{All} & \textsc{Freq} & \textsc{Rare}\\
\midrule
$L_1$: \xmark; $\mathcal{L}_{iou}$:\cmark& 34.9& 56.4& 27.3& 75.2& 82.1 & 53.8\\
$L_1$: \cmark; $\mathcal{L}_{iou}$:\xmark& 37.7& 57.8& 28.4&76.0&82.7&54.6\\
$L_1$: \cmark; $\mathcal{L}_{iou}$:\cmark& \textbf{39.1}&\textbf{58.1}&\textbf{29.1}& \textbf{76.8}&\textbf{83.3}&\textbf{55.1}\\
\bottomrule
\end{tabular}}
\label{tab:ablation_loss_ek}
\end{table}
\noindent
\textbf{Ablation: Loss function.} The training loss function defined in Eq. (4) in the main paper contains three components (cross-entropy loss and two temporal losses). We conduct ablation experiments by removing one of the temporal losses. Note that we always need cross entropy loss for the classification task. Results in Table~\ref{tab:ablation_loss} and Table~\ref{tab:ablation_loss_ek} show that models trained with overall loss perform better than the ones trained with the ablated versions. Moreover, the models trained with only $L_1$ temporal loss perform better than the ones trained with only $\mathcal{L}_{iou}$.

\noindent 
\textbf{Ablation: Anticipation queries.} The number of anticipation queries discerns the maximum number of action instances the model is supposed to predict. Results in Table~\ref{tab:ablation_nqueries} and Table~\ref{tab:ablation_nqueries_ek} shows the performance of our model with different number of anticipation queries. The results suggest minor improvement with higher number of anticipation queries, however, the models with more number of queries require longer training times. Intuitively, a very large number of anticipation queries implies the model will require more time to learn the non-maximal suppression of the irrelevant predictions. On the other hand, when the number of anticipation queries is reduced, the anticipation performance of our model degrades. A very small number of anticipation queries implies less number of action are anticipated. Thus, for very complex video with many future action instances, the model would miss several action instances resulting in poor anticipation performance. Additionally, as shown in Table~\ref{tab:ablation_nqueries}, the anticipation error increases over time. This is because there are more actions to be anticipated and the model is limited by the number of anticipation queries.

\setlength{\tabcolsep}{4pt}
\renewcommand{\arraystretch}{0.95}
\begin{table}[t]
\centering
\caption{\textbf{Ablation: Anticipation Queries (Breakfast and 50Salads).} We report the mean over classes accuracy for different observation/anticipation durations. Higher values indicate better performance.}
\scalebox{0.54}{
\begin{tabular}{lll@{\hskip 5mm}c@{\hskip 5mm}c@{\hskip 5mm}c@{\hskip 5mm}c@{\hskip 5mm}c@{\hskip 5mm}c@{\hskip 5mm}c@{\hskip 5mm}c}
\toprule
& Method & $\beta_o$\ $\rightarrow$ & \multicolumn{4}{c}{20\%} & \multicolumn{4}{c}{30\%} \\
\cmidrule(lr){4-7}\cmidrule(lr){8-11}
& & $\beta_a$\ $\rightarrow$ & 10\% & 20\% & 30\% & 50\% & 10\% & 20\% & 30\% & 50\%\\
\midrule 
\textbf{Breakfast}
&$N_a = 50$&& 32.6& 28.2& 26.4& 24.3&35.8& 31.4& 28.7& 25.3\\
&$N_a = 150$&& \textbf{37.4}&\textbf{32.0}&\textbf{30.3}&\textbf{28.6}&\textbf{39.9}& \textbf{35.7}&\textbf{32.1}&\textbf{29.4}\\
&$N_a = 500$&& 36.6& 31.5& 29.4& 27.3&38.5& 34.4&31.3&28.3\\
\midrule
\textbf{50Salads}
&$N_a = 20$&& 38.4& 33.2& 24.2& 23.6& 39.1& 35.6& 25.5&24.2\\
&$N_a = 80$&&\textbf{41.1}&\textbf{35.0}&\textbf{27.6}&\textbf{27.3}&\textbf{42.8}&\textbf{42.3}&\textbf{28.5}&\textbf{23.6}\\
&$N_a = 320$&& 40.5& 34.2& 26.0& 25.6& 41.3& 40.9&27.4&23.3\\
\bottomrule
\end{tabular}}
\label{tab:ablation_nqueries}
\end{table}

\setlength{\tabcolsep}{4pt}
\renewcommand{\arraystretch}{0.98}
\begin{table}[t]
\centering
\caption{\textbf{Ablation: Anticipation Queries (EK-55 and EGTEA+).} We report mAP values for \textsc{all} classes, \textsc{frequent} classes ($> 100$ action instances) and \textsc{rare} class ($< 10$ action instances). Following ~\cite{nagarajan2020ego}, we report the mAP values averaged over different observation durations. Higher values implies better performance.}
\scalebox{0.7}{
\begin{tabular}{l@{\hskip 3mm}l@{\hskip 3mm}ccc}
\toprule
Dataset& & \textsc{All} & \textsc{Freq} & \textsc{Rare}\\
\midrule
\textbf{EK-55}&$N_a = 300$& 34.3& 55.6& 24.2\\
&$N_a = 900$ & \textbf{39.1}&\textbf{58.1}&\textbf{29.1}\\
&$N_a = 2700$& 38.2&56.9&28.3\\
\midrule
\textbf{EGTEA+} & $N_a = 200$& 70.2& 79.5& 49.7\\
&$N_a = 600$ & \textbf{76.8}&\textbf{83.3}&\textbf{55.1}\\
&$N_a = 1800$& 75.3& 82.4& 53.3\\
\bottomrule
\end{tabular}}
\label{tab:ablation_nqueries_ek}
\end{table}
\noindent 
\textbf{Ablation: Segment window length.} Results in Table~\ref{tab:ablation_windowsize} and Table~\ref{tab:ablation_windowsize_ek} shows the performance of our model with different values of temporal window lengths used to extract segment-level representations during action anticipation. The results suggest that neither a very small window length nor a very large window is helpful. The segment encoder is trained to predict future actions given a video segment depicting a single action. During the action anticipation stage, when the segment encoder is used to extract segment-level representations, the observed video is divided into a series of non-overlapping segment using temporal sliding windows as the action boundaries are not known. Intuitively, when the temporal sliding window is very small, the individual segments do not have enough information to obtain effective representations. On the other hand, when the window is very large, the segments contain more than one action and potentially results in segment-level representations with overlapping semantic content. We observe that the drop in performance with models that use smaller window lengths is larger as compared to the ones with larger window lengths.
\setlength{\tabcolsep}{4pt}
\renewcommand{\arraystretch}{0.95}
\begin{table}[t]
\centering
\caption{\textbf{Ablation: Segment window length (Breakfast and 50Salads).} We report the mean over classes accuracy for different observation/anticipation durations. Higher values indicate better performance. }
\scalebox{0.54}{
\begin{tabular}{lll@{\hskip 5mm}c@{\hskip 5mm}c@{\hskip 5mm}c@{\hskip 5mm}c@{\hskip 5mm}c@{\hskip 5mm}c@{\hskip 5mm}c@{\hskip 5mm}c}
\toprule
& Method & $\beta_o$\ $\rightarrow$ & \multicolumn{4}{c}{20\%} & \multicolumn{4}{c}{30\%} \\
\cmidrule(lr){4-7}\cmidrule(lr){8-11}
& & $\beta_a$\ $\rightarrow$ & 10\% & 20\% & 30\% & 50\% & 10\% & 20\% & 30\% & 50\%\\
\midrule 
\textbf{Breakfast}
&$k = 4$&& 35.9& 30.6& 26.3& 26.1& 38.4& 33.6&30.8&28.2\\
&$k = 16$&& \textbf{37.4}&\textbf{32.0}&\textbf{30.3}&\textbf{28.6}&\textbf{39.9}& \textbf{35.7}&\textbf{32.1}&\textbf{29.4}\\
&$k = 64$&& 37.4& 31.7& 29.9& 28.1& 39.1& 35.0&31.7&28.7\\
\midrule
\textbf{50Salads}
&$k = 12$&& 39.0& 33.5& 25.8& 25.4& 39.6& 38.4&26.4&21.5\\
&$k = 48$&&\textbf{41.1}&\textbf{35.0}&\textbf{27.6}&\textbf{27.3}&\textbf{42.8}&\textbf{42.3}&\textbf{28.5}&\textbf{23.6}\\
&$k = 192$&& 41.0& 34.8& 27.2& 26.8& 42.6& 42.1&27.5&22.8\\
\bottomrule
\end{tabular}}
\label{tab:ablation_windowsize}
\end{table}

\setlength{\tabcolsep}{4pt}
\renewcommand{\arraystretch}{0.98}
\begin{table}[t]
\centering
\caption{\textbf{Ablation: Segment window length (EK-55 and EGTEA+).} We report mAP values for \textsc{all} classes, \textsc{frequent} classes ($> 100$ action instances) and \textsc{rare} class ($< 10$ action instances). Following ~\cite{nagarajan2020ego}, we report the mAP values averaged over different observation durations. Higher values implies better performance.}
\scalebox{0.7}{
\begin{tabular}{l@{\hskip 3mm}l@{\hskip 3mm}ccc}
\toprule
Dataset& & \textsc{All} & \textsc{Freq} & \textsc{Rare}\\
\midrule
\textbf{EK-55}&$k = 8$& 37.9& 57.2& 27.4\\
&$k = 32$ & \textbf{39.1}&\textbf{58.1}&\textbf{29.1}\\
&$k = 128$& 38.8& 58.0& 28.7\\
\midrule
\textbf{EGTEA+} & $k = 6$& 75.4& 81.7& 53.9\\
&$k = 24$ & \textbf{76.8}&\textbf{83.3}&\textbf{55.1}\\
&$k = 96$& 76.3& 82.9& 54.8\\
\bottomrule
\end{tabular}}
\label{tab:ablation_windowsize_ek}
\end{table}

\noindent
\textbf{Ablation: Sliding Windows for Segment Encoder Training.} 
Instead of using action boundaries we used sliding temporal windows of length=$k$ (same as used during stage 2) to obtain segments for segment-level training. Results in Table~\ref{tab:ablation_sw} and Table~\ref{tab:ablation_sw_ek} show that this approach results in a slightly lower performance than our proposed training approach. This is possibly due to increased noise in the segment-level representations from this training approach.
\setlength{\tabcolsep}{4pt}
\renewcommand{\arraystretch}{1.0}
\begin{table}[h]
\centering
\caption{\textbf{Ablation: Sliding windows for Segment Encoder Training.} Mean over classes accuracy for different observation/anticipation durations. Higher is better. [BF: Breakfast; 50SL: 50Salads]}
\scalebox{0.65}{
\begin{tabular}{ll@{\hskip 5mm}c@{\hskip 5mm}c@{\hskip 5mm}c@{\hskip 5mm}c@{\hskip 5mm}c@{\hskip 5mm}c@{\hskip 5mm}c@{\hskip 5mm}c}
\toprule
& Observation ($\beta_o$)\ $\rightarrow$ & \multicolumn{4}{c}{20\%} & \multicolumn{4}{c}{30\%} \\
\cmidrule(lr){3-6}\cmidrule(lr){7-10}
& Anticipation ($\beta_a$)\ $\rightarrow$ & 10\% & 20\% & 30\% & 50\% & 10\% & 20\% & 30\% & 50\%\\
\midrule 
\textbf{Breakfast}& Sliding windows&35.9&30.7&28.0&26.4&37.8&33.5&29.9&25.2\\
& \textbf{\textsc{Anticipatr}(Full)} & \textbf{37.4}&\textbf{32.0}&\textbf{30.3}&\textbf{28.6}&\textbf{39.9}& \textbf{35.7}&\textbf{32.1}&\textbf{29.4}\\

\midrule
\textbf{50Salads}&Sliding windows&37.2&33.5&26.3&25.8&37.9&37.0&26.1&24.5\\
& \textbf{\textsc{Anticipatr}(Full)} &\textbf{41.1}&\textbf{35.0}&\textbf{27.6}&\textbf{27.3}&\textbf{42.8}&\textbf{42.3}&\textbf{28.5}&\textbf{23.6}\\
\bottomrule
\end{tabular}}
\vspace{-3mm}
\label{tab:ablation_sw}
\end{table}

\setlength{\tabcolsep}{4pt}
\renewcommand{\arraystretch}{0.98}
\begin{table}[h]
\centering
\caption{\textbf{Ablation: Sliding windows for Segment Encoder Training.} mAP values for \textsc{all} classes, \textsc{frequent} classes ($> 100$ action instances) and \textsc{rare} class ($< 10$ action instances). Higher is better.}
\scalebox{0.85}{
\begin{tabular}{l@{\hskip 3mm}ccc@{\hskip 4mm}ccc}
\toprule
Method & \multicolumn{3}{c}{\textbf{EK-55}} & \multicolumn{3}{c}{\textbf{EGTEA+}}\\
\cmidrule(lr){2-4}\cmidrule(lr){5-7}
& \textsc{All} & \textsc{Freq} & \textsc{Rare} & \textsc{All} & \textsc{Freq} & \textsc{Rare}\\
\midrule

Sliding Windows & 37.6 & 56.5 & 27.4 & 74.8 & 81.2 & 53.0\\
No Video Encoder & 30.9 & 51.8 & 21.2& 70.2 & 79.9 & 50.1\\
\textbf{\textsc{Anticipatr}(Full)}& \textbf{39.1}&\textbf{58.1}&\textbf{29.1}& \textbf{76.8}&\textbf{83.3}&\textbf{55.1}\\
\bottomrule
\end{tabular}}
\label{tab:ablation_sw_ek}
\end{table}





\noindent 
\textbf{Ablation: Set correspondence.} To compute the anticipation loss, we use a greedy algorithm to align groundtruth and predicted set of action instances. Another commonly employed set correspondence algorithm is Hungarian matcher algorithm used in prior works~\cite{carion2020end,kim2021hotr}. For completeness, we also conducted experiments with Hungarian matcher optimized over the cost function with all three terms (classification loss and two temporal losses) following ~\cite{nawhal2021activity}. We didn't observe any significant difference in performance of the models trained using either of the two matchers as shown in Table~\ref{tab:ablation_matcher} and Table~\ref{tab:ablation_matcher_ek}.

\setlength{\tabcolsep}{4pt}
\renewcommand{\arraystretch}{0.95}
\begin{table}[t]
\centering
\caption{\textbf{Ablation: Set correspondence (Breakfast \& 50Salads).} We report the mean over classes accuracy for different observation/anticipation durations. Higher values indicate better performance.}
\scalebox{0.54}{
\begin{tabular}{lll@{\hskip 5mm}c@{\hskip 5mm}c@{\hskip 5mm}c@{\hskip 5mm}c@{\hskip 5mm}c@{\hskip 5mm}c@{\hskip 5mm}c@{\hskip 5mm}c}
\toprule
& Method & $\beta_o$\ $\rightarrow$ & \multicolumn{4}{c}{20\%} & \multicolumn{4}{c}{30\%} \\
\cmidrule(lr){4-7}\cmidrule(lr){8-11}
& & $\beta_a$\ $\rightarrow$ & 10\% & 20\% & 30\% & 50\% & 10\% & 20\% & 30\% & 50\%\\
\midrule 
\textbf{Breakfast}
&Hungarian&& 36.8 & \textbf{32.0}& \textbf{30.5}& 28.4& 39.2 & 35.4& 31.9& \textbf{29.6}\\
&Greedy && \textbf{37.4}&32.0&\textbf{30.3}&\textbf{28.6}&\textbf{39.9}& \textbf{35.7}&\textbf{32.1}&29.4\\
\midrule
\textbf{50Salads}
&Hungarian && \textbf{41.3}& \textbf{35.1}& 27.4& 26.8& \textbf{42.9}& 42.0& 28.4& \textbf{23.8}\\
&Greedy &&41.1&35.0&\textbf{27.6}&\textbf{27.3}&\textbf{42.8}&\textbf{42.3}&\textbf{28.5}&\textbf{23.6}\\
\bottomrule
\end{tabular}}
\label{tab:ablation_matcher}
\end{table}

\setlength{\tabcolsep}{4pt}
\renewcommand{\arraystretch}{0.98}
\begin{table}[t]
\centering
\caption{\textbf{Ablation: Set correspondence
(EK-55 \& EGTEA+).} We report mAP values for \textsc{all} classes, \textsc{frequent} classes ($> 100$ action instances) and \textsc{rare} class ($< 10$ action instances). Following ~\cite{nagarajan2020ego}, we report the mAP values averaged over different observation durations. Higher values implies better performance.}
\scalebox{0.8}{
\begin{tabular}{l@{\hskip 3mm}ccc@{\hskip 4mm}ccc}
\toprule
Method & \multicolumn{3}{c}{\textbf{EK-55}} & \multicolumn{3}{c}{\textbf{EGTEA+}}\\
\cmidrule(lr){2-4}\cmidrule(lr){5-7}
& \textsc{All} & \textsc{Freq} & \textsc{Rare} & \textsc{All} & \textsc{Freq} & \textsc{Rare}\\
\midrule
Hungarian&39.0&\textbf{58.4}&28.4&76.7&\textbf{83.5}&55.0\\
Greedy& \textbf{39.1}&58.1&\textbf{29.1}& \textbf{76.8}&83.3&\textbf{55.1}\\
\bottomrule
\end{tabular}}
\label{tab:ablation_matcher_ek}
\end{table}

\subsection{Additional Qualitative Analysis}
\label{sec:qual}
\begin{figure}[h]
    \centering
    \begin{tabular}{cc}
    \includegraphics[width=0.49\textwidth]{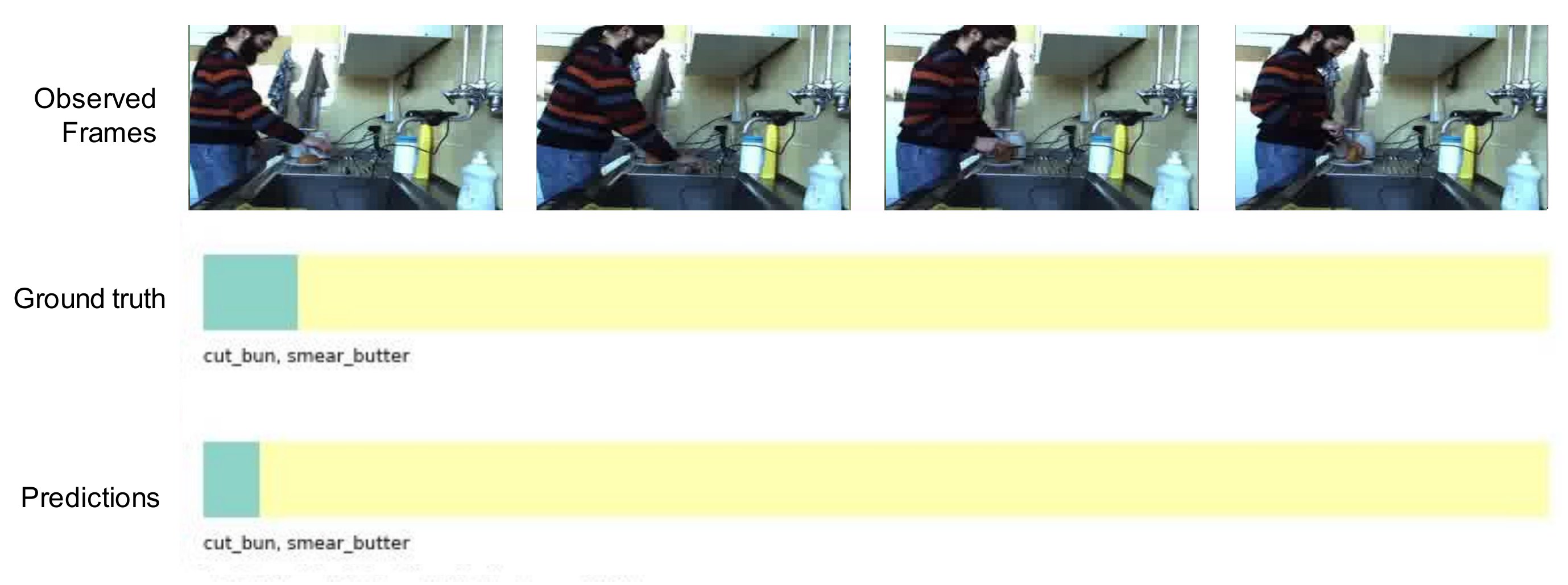}&
    \includegraphics[width=0.49\textwidth]{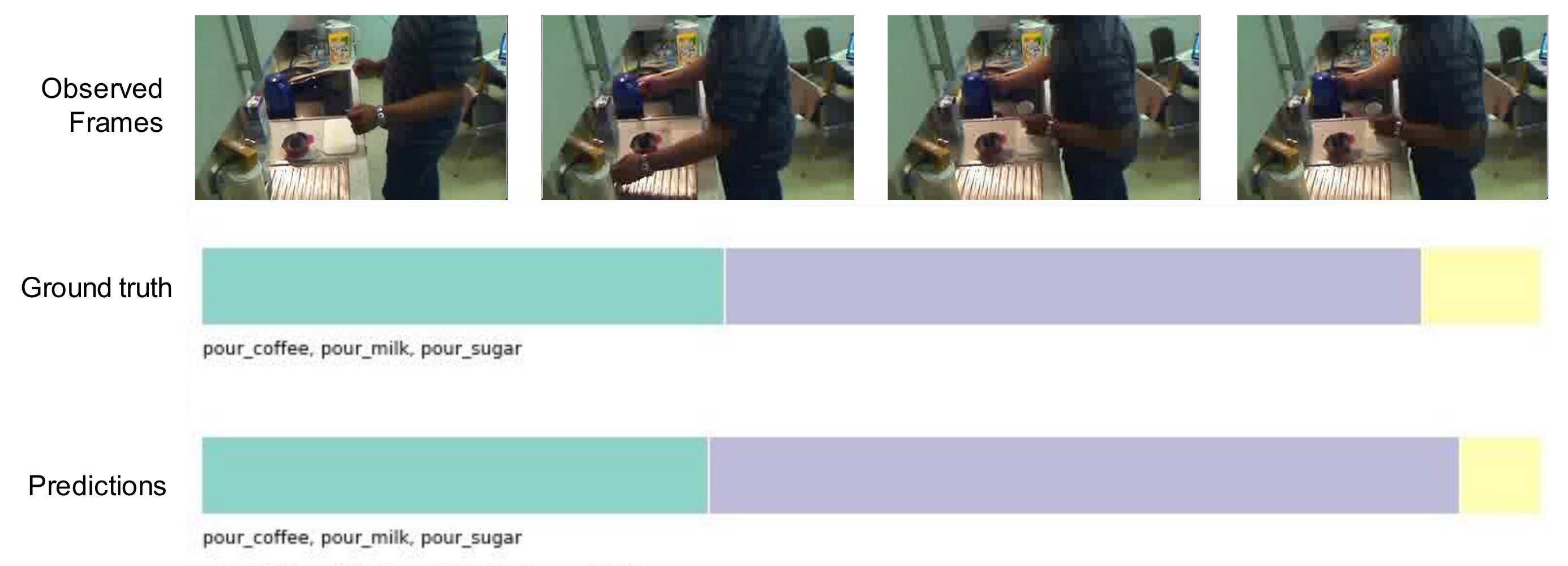}
    \\
    (a) & (b)\\
    \includegraphics[width=0.49\textwidth]{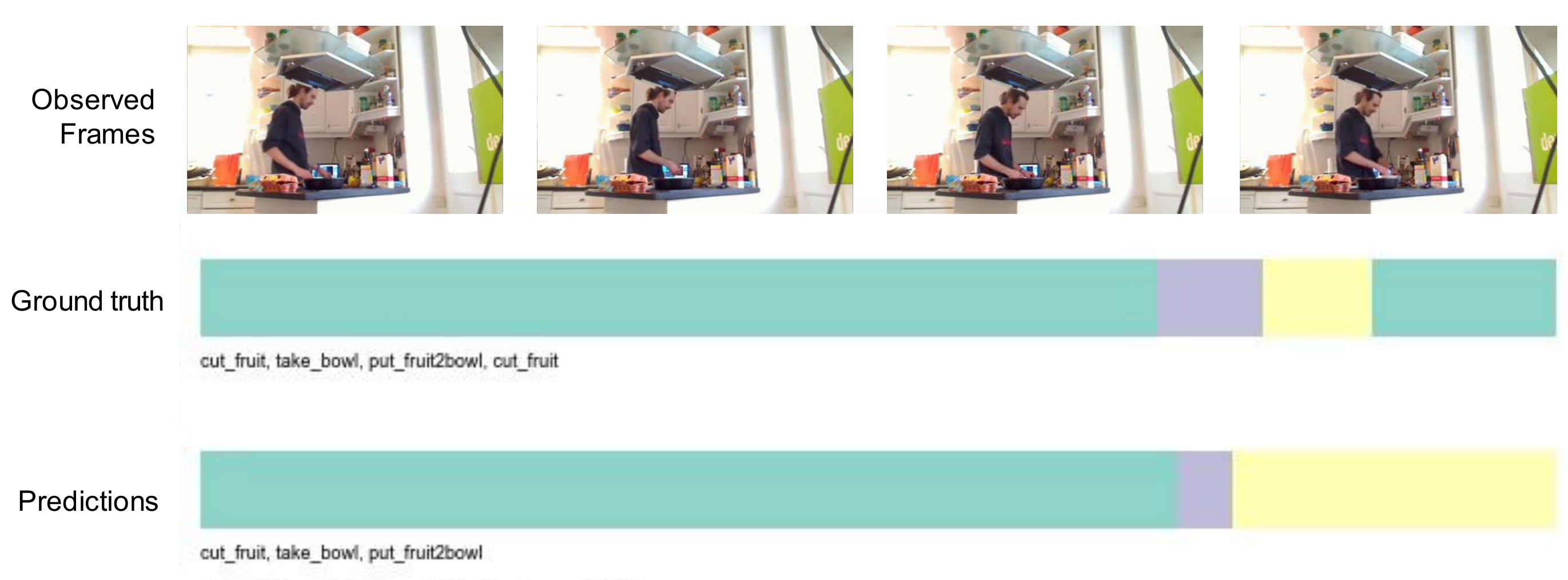}&
    \includegraphics[width=0.49\textwidth]{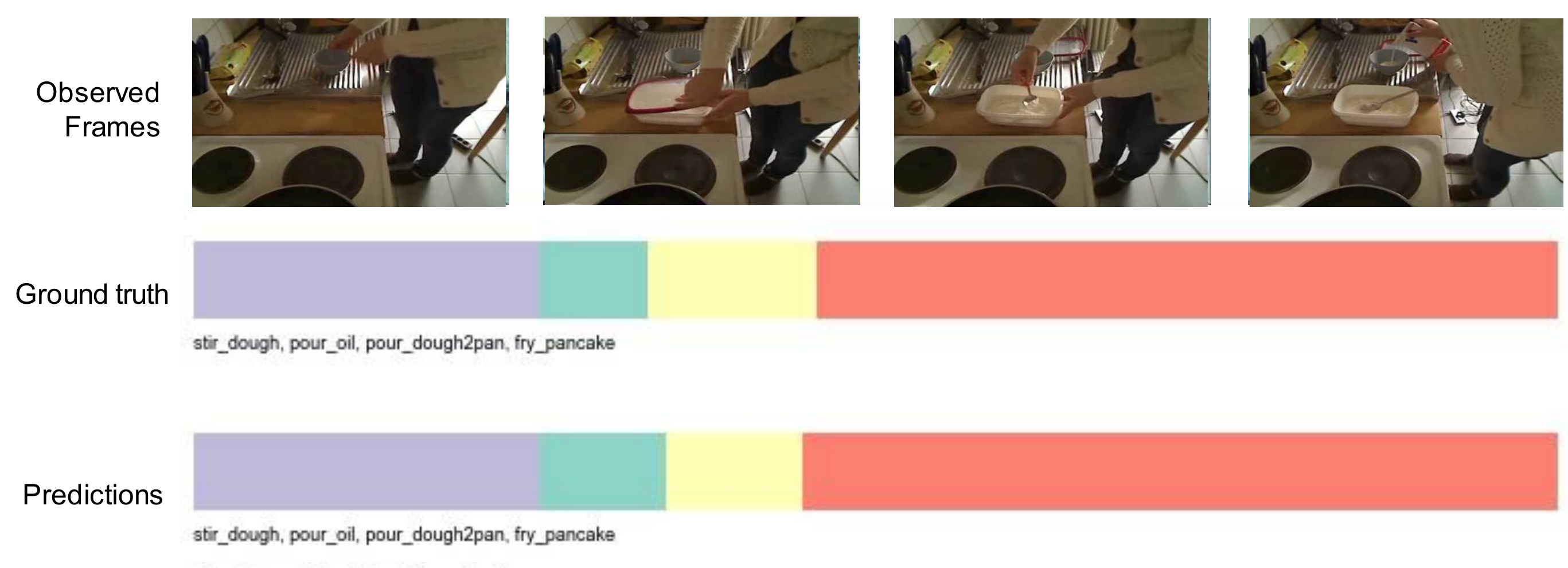}\\
    (c) & (d)\\
    \end{tabular}
\caption{\textbf{Visualizations (Breakfast).} Examples from Breakfast dataset for the case where observation duration is 20\% of the video duration and anticipation duration involves predicting actions for 50\% of the remaining video.}
\label{fig:qual_ltaa_bf}
\end{figure}

\begin{figure}[t]
    \centering
    \begin{tabular}{cc}
    \includegraphics[width=0.49\textwidth]{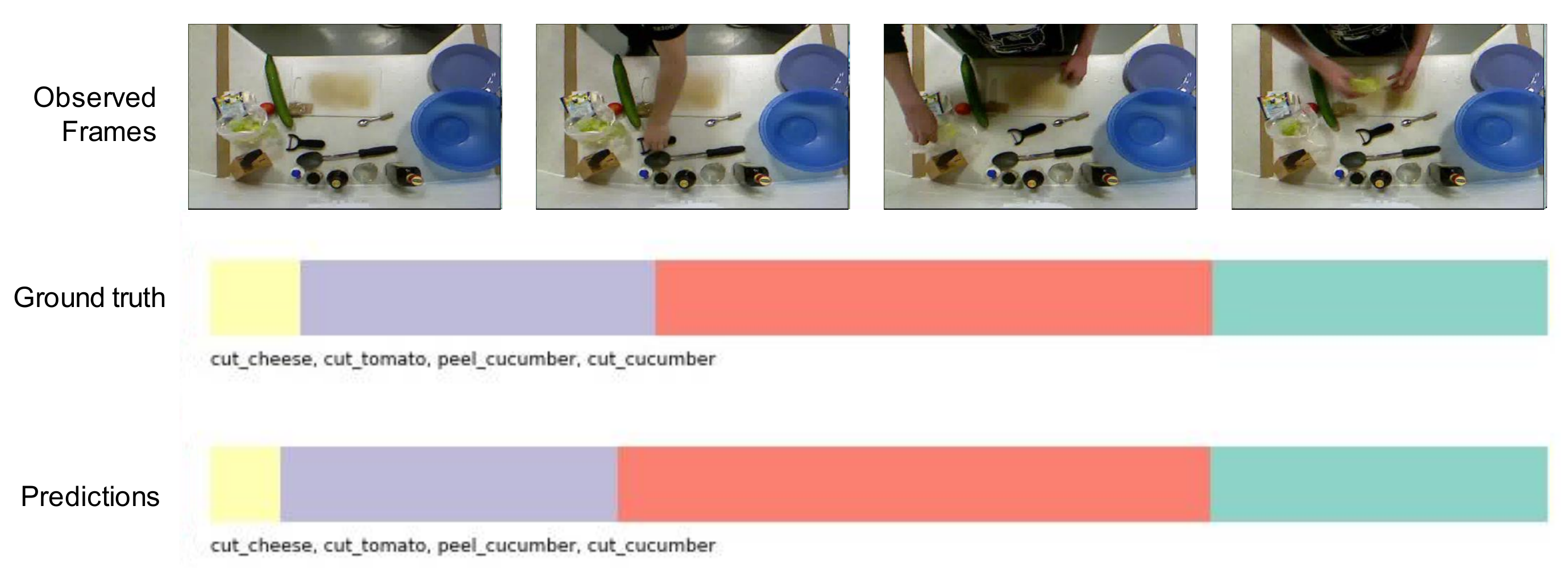}&
    \includegraphics[width=0.49\textwidth]{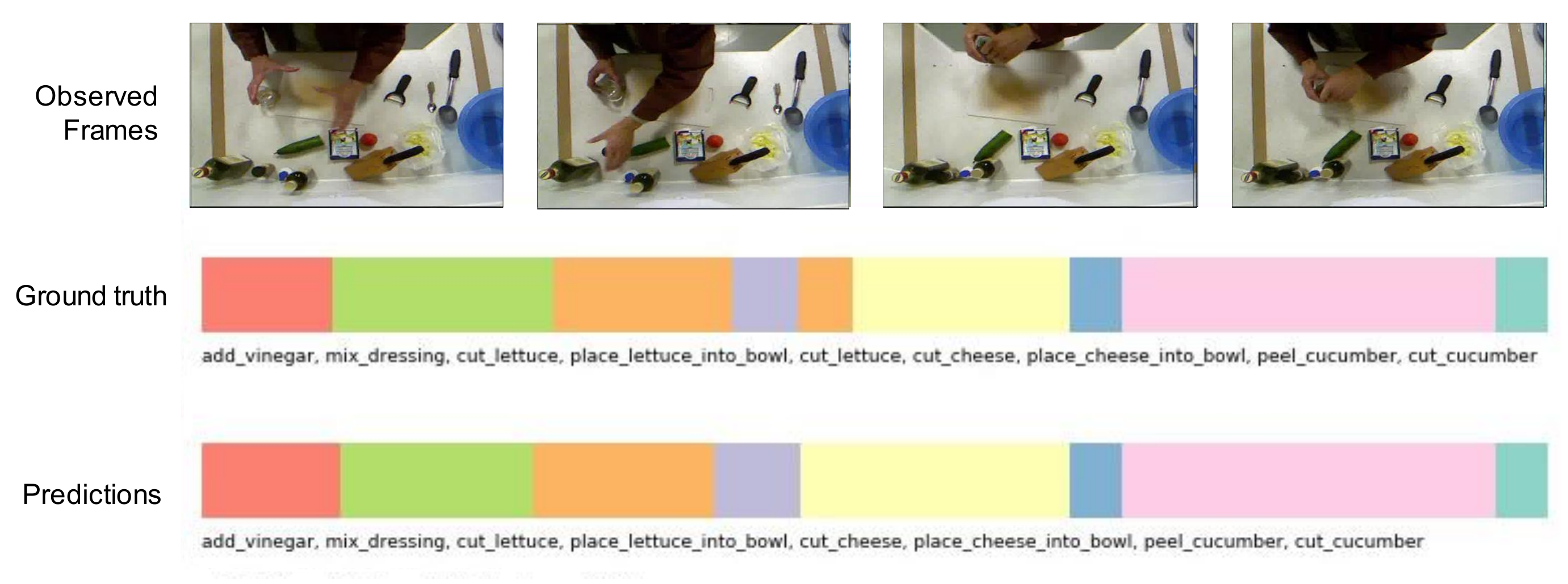}\\
    (a) & (b)\\
    \includegraphics[width=0.49\textwidth]{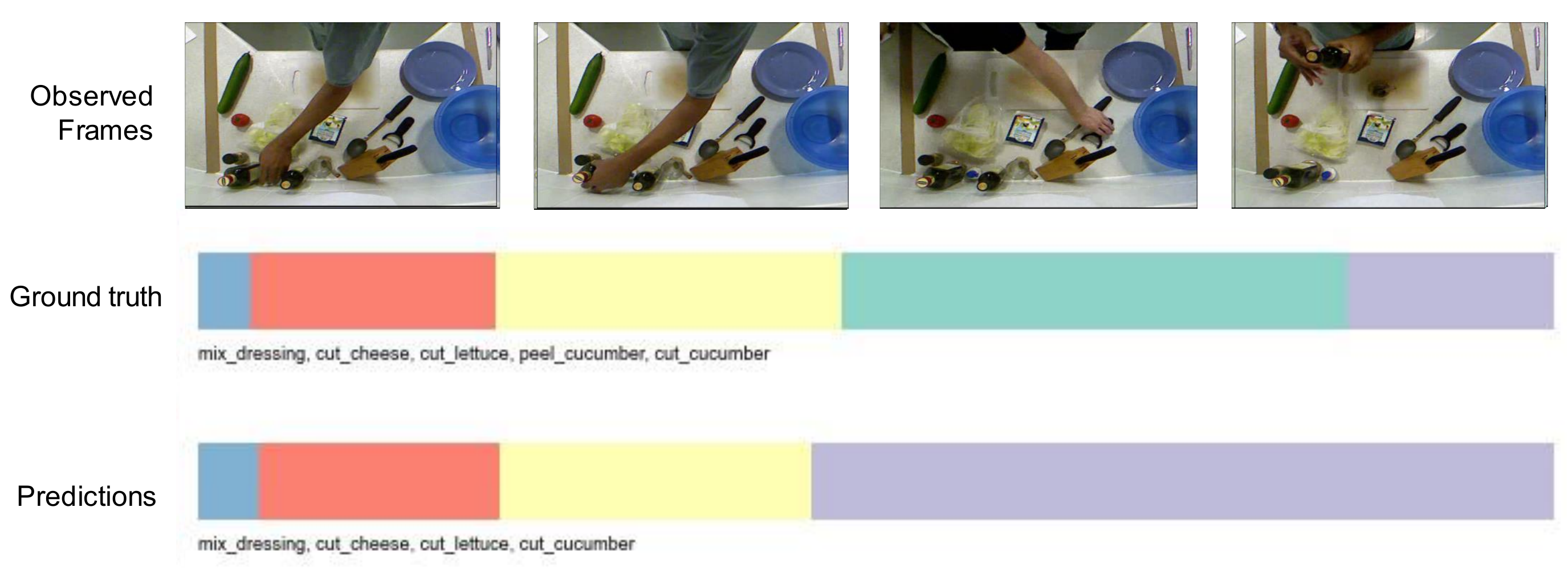}&
    \includegraphics[width=0.49\textwidth]{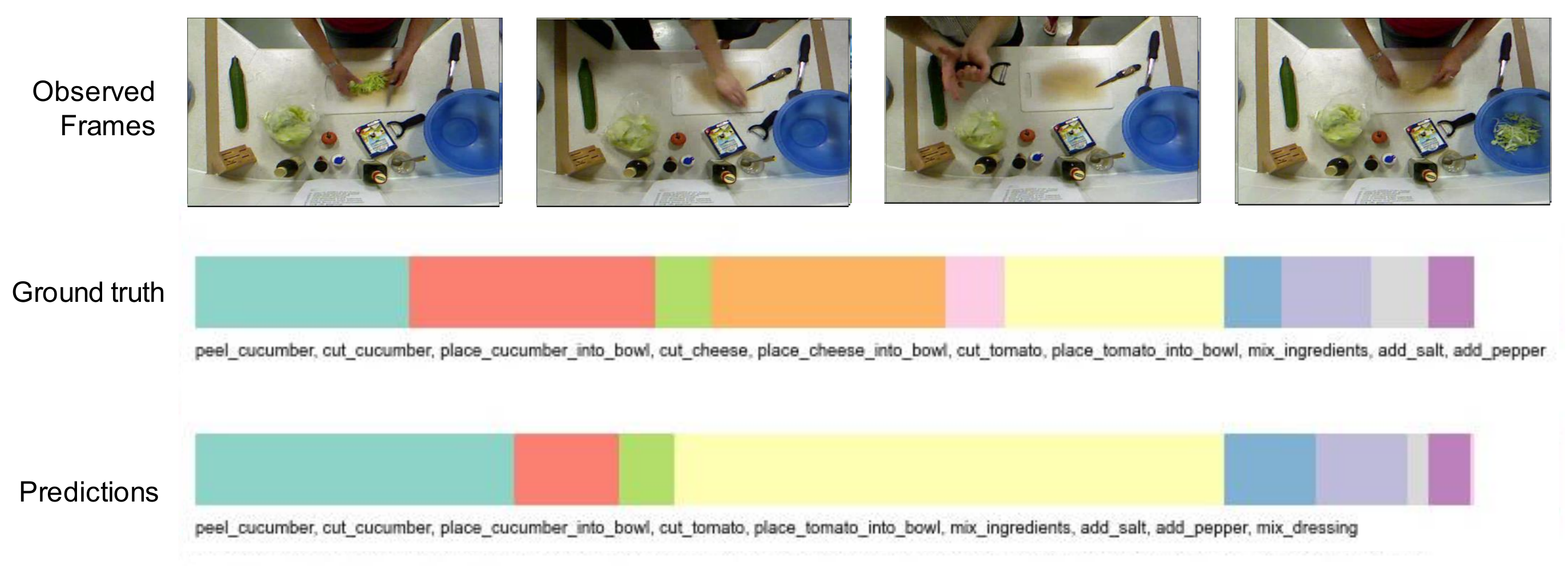}\\
    (c) & (d)\\
    \end{tabular}
\caption{\textbf{Visualizations (50Salads).} Examples from 50Salads dataset for the case where observation duration is 20\% of the video duration and anticipation duration involves predicting actions for 50\% of the remaining video.}
\label{fig:qual_ltaa_salads}
\end{figure}

Visualizations in Fig.~\ref{fig:qual_ltaa_bf} and Fig.~\ref{fig:qual_ltaa_salads} show that our model is generally able to anticipate correct actions at future time instants long anticipation durations for Breakfast and 50Salads benchmarks respectively.

\begin{figure}[t]
    \centering
    \begin{tabular}{c}
    \includegraphics[width=0.55\textwidth]{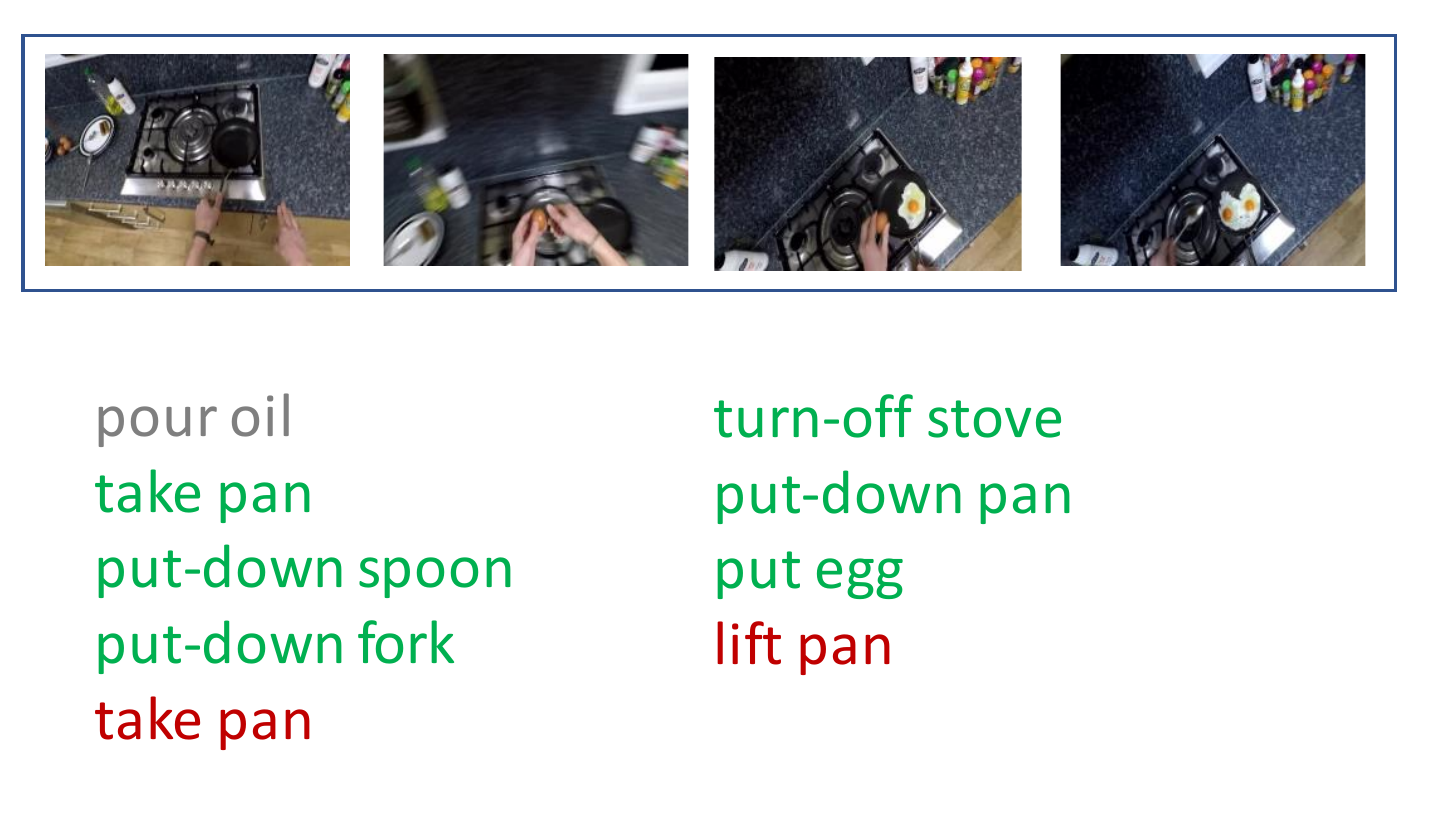}\\
    (a)\\
    \includegraphics[width=0.55\textwidth]{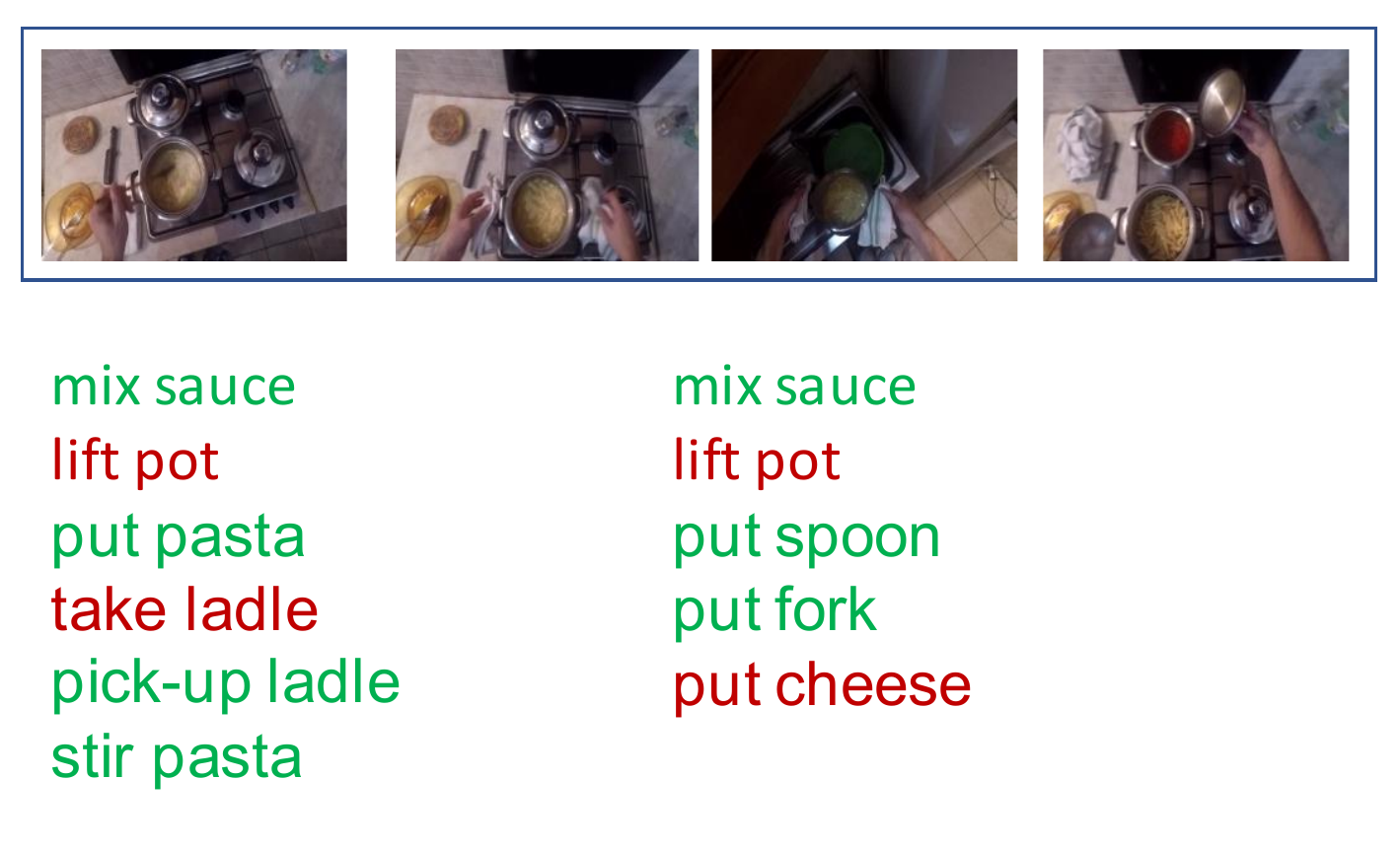}\\
    (b)\\
    \includegraphics[width=0.55\textwidth]{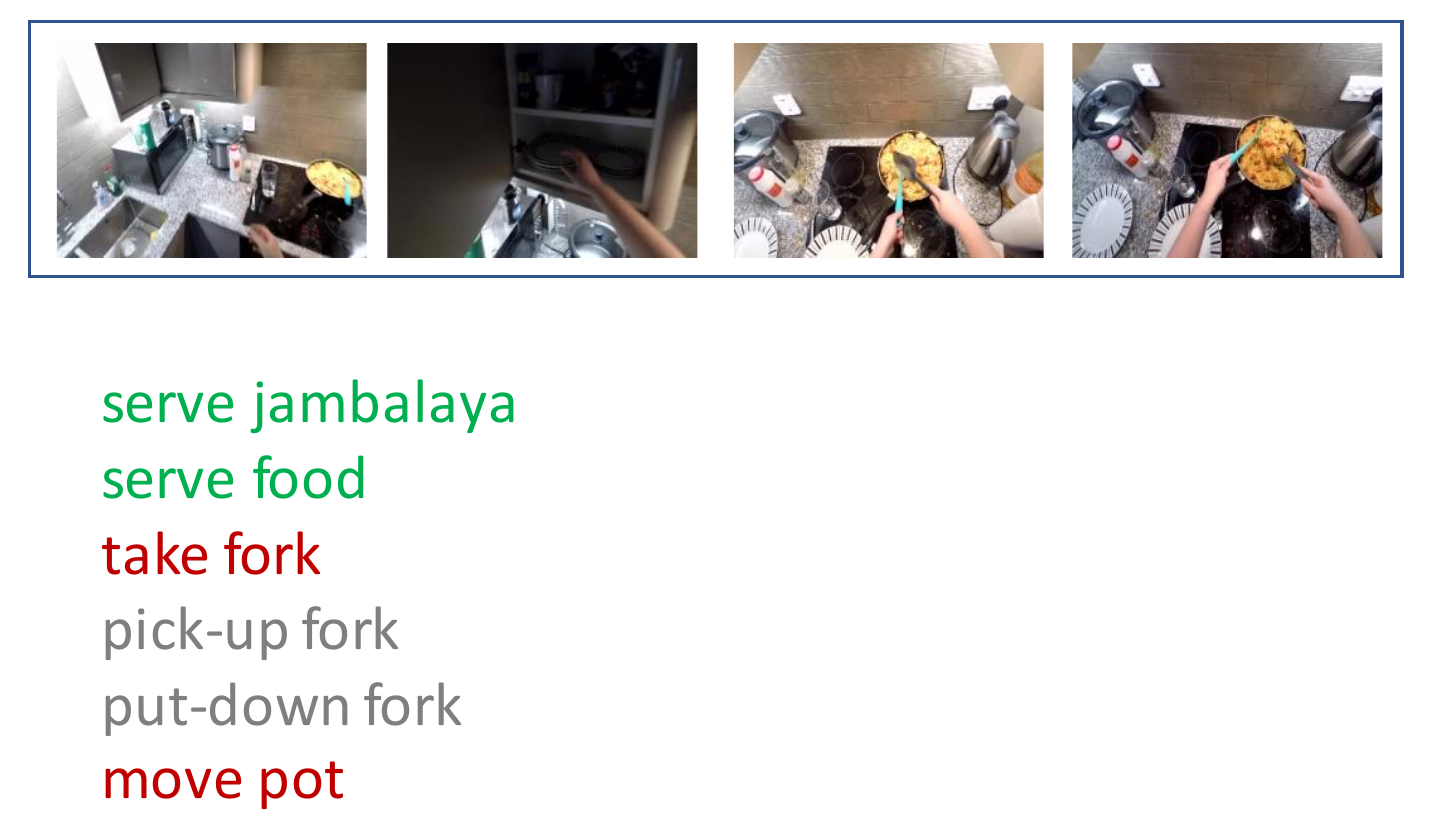}\\
    (c)\\
    \includegraphics[width=0.55\textwidth]{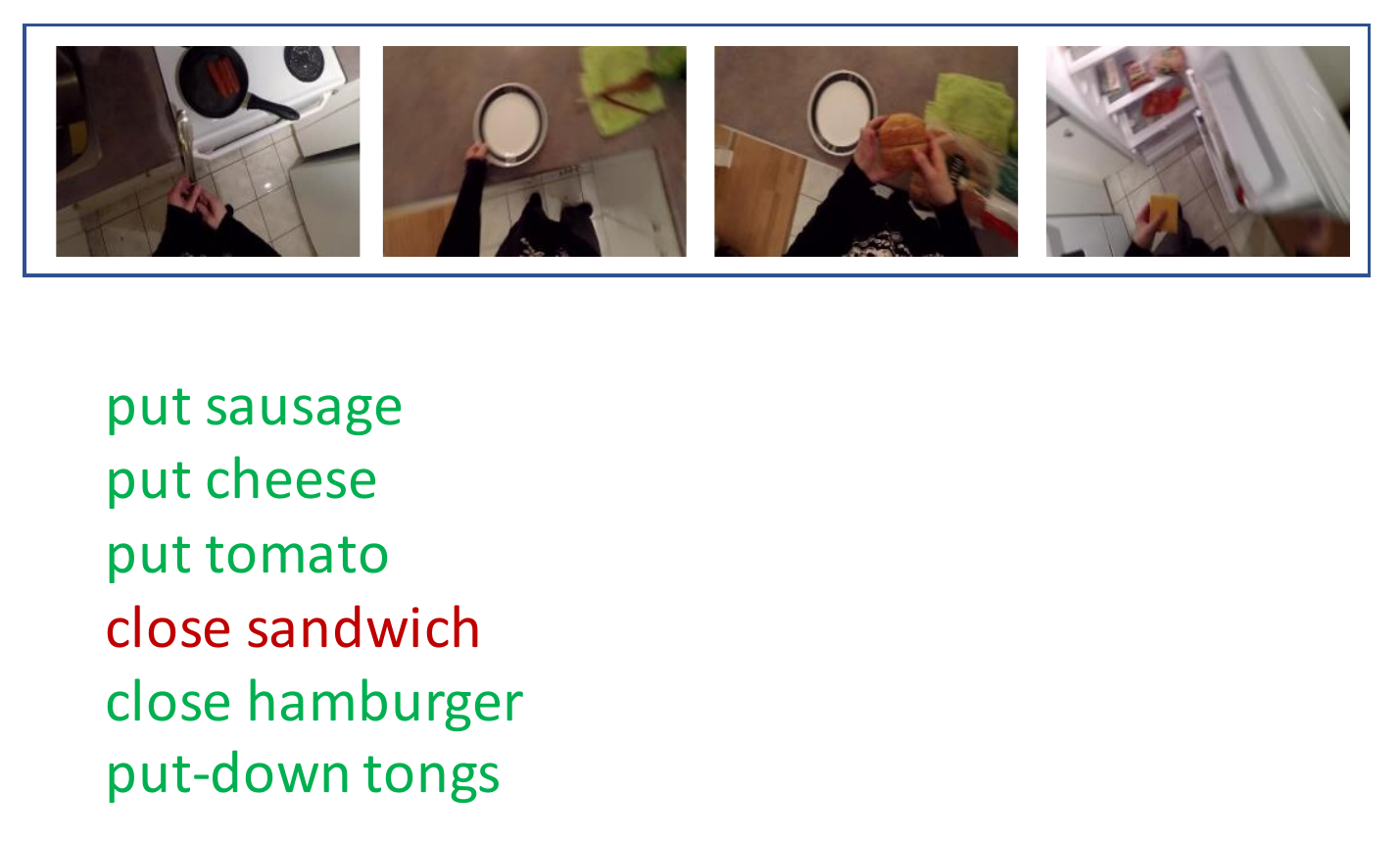}\\
    (d)
    \end{tabular}
    \caption{\textbf{Visualizations (EK-55).} Examples from Epic-Kitchens-55 dataset for the case where observation duration is 50\% of the video duration. We show the predicted action classes in the visualization -- classes in green color are correct predictions, classes in red color are wrong predictions, and classes in gray color are missed classes.}
    \label{fig:qual_ltaa_ek}
\end{figure}

\begin{figure}[t]
    \centering
    \begin{tabular}{c}
    \includegraphics[width=0.6\textwidth]{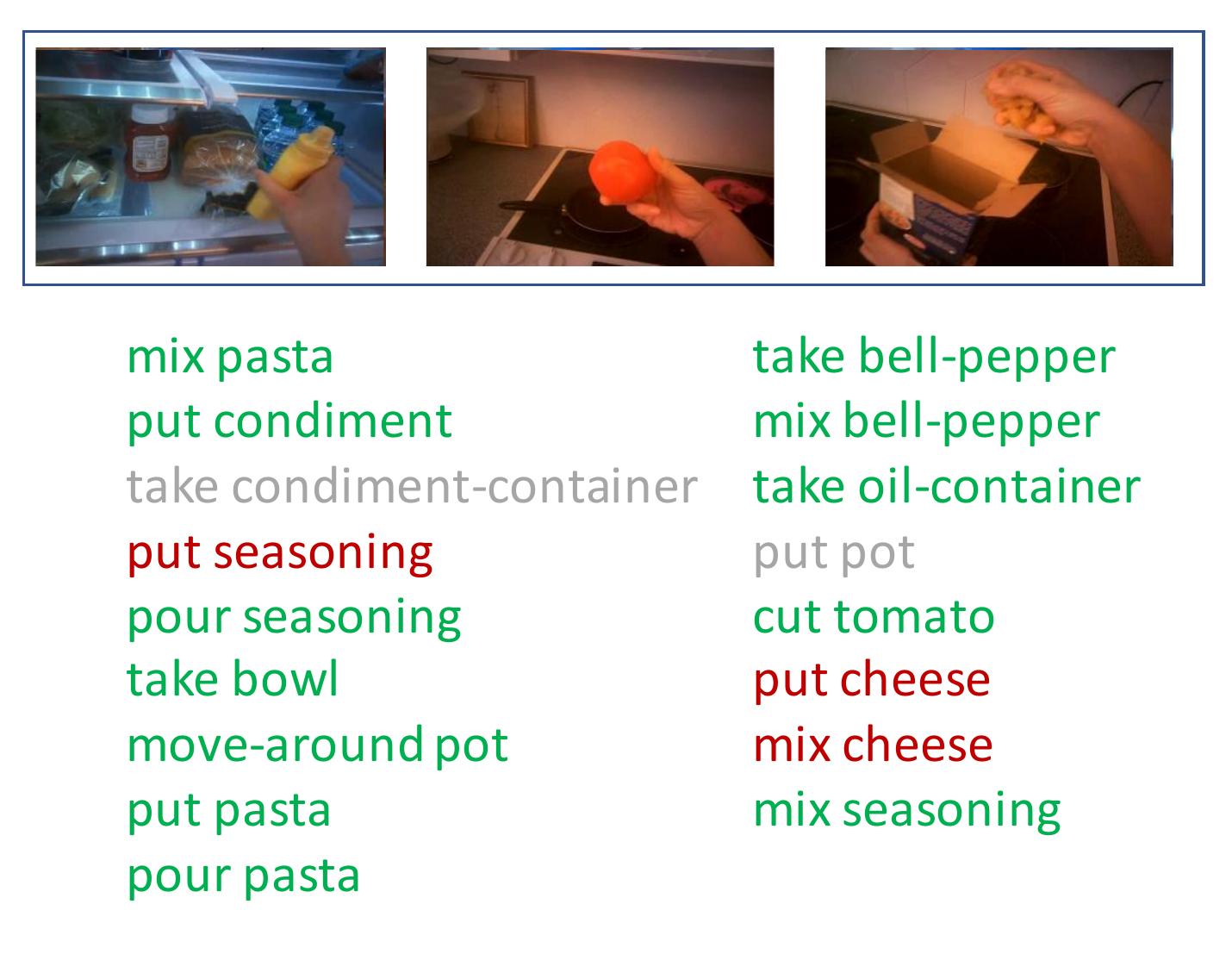}
    \\
    (a)\\
    \includegraphics[width=0.6\textwidth]{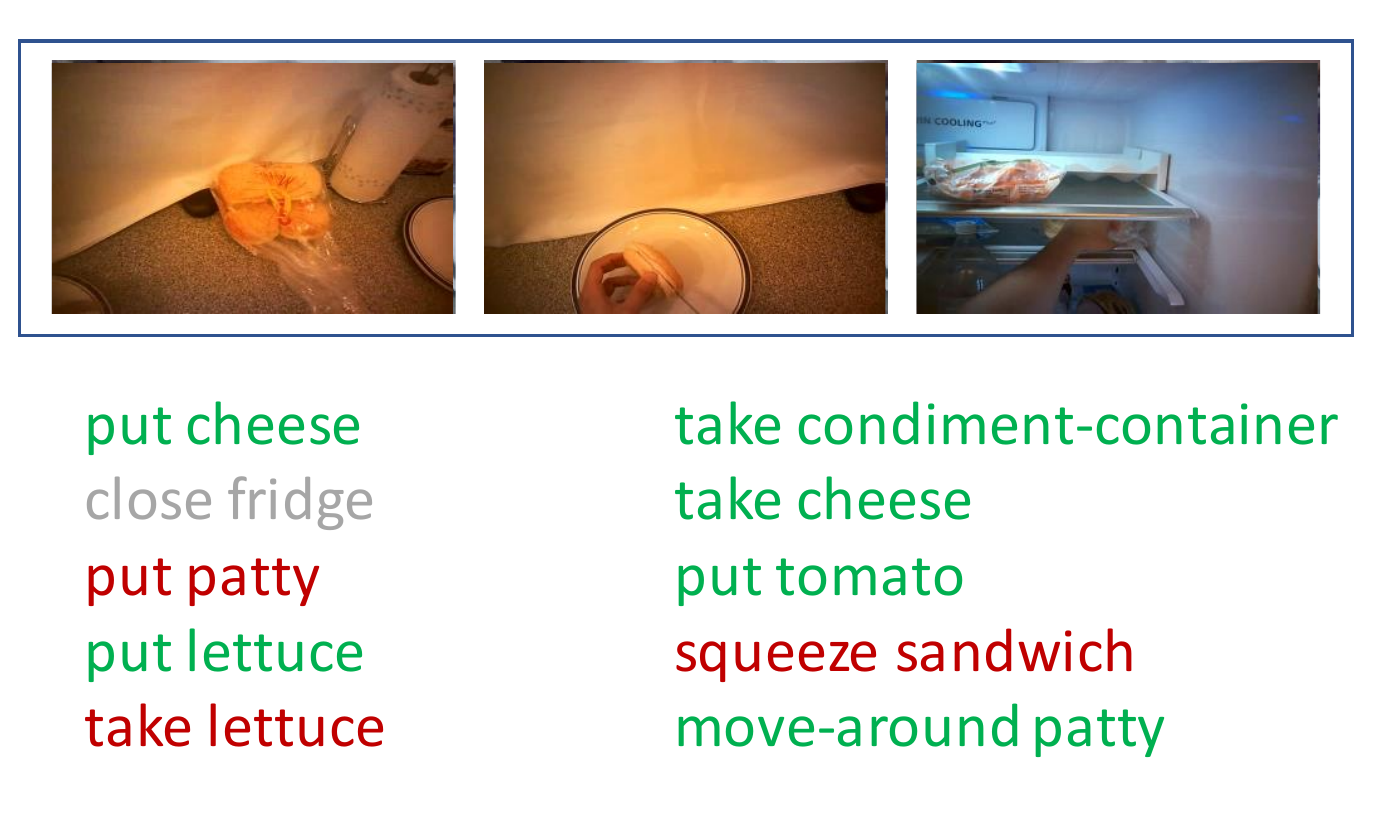}\\
    (b)
    \end{tabular}
    \caption{\textbf{Visualizations (EGTEA+).} Examples from EGTEA Gaze+ dataset for the case where 50\% of the video is observed. We show the predicted action classes in the visualization -- classes in green color are correct predictions, classes in red color are wrong predictions, and classes in gray color are missed classes.}
    \label{fig:qual_ltaa_egtea}
\end{figure}

Visualizations in Fig.~\ref{fig:qual_ltaa_ek} and Fig.~\ref{fig:qual_ltaa_egtea} show that our model is able to effectively predict future action classes for EK-55 and EGTEA benchmarks respectively.

\noindent
\textbf{Failure Cases.} We observe that the action boundaries in some cases are not exactly aligned with the groundtruth even though the class labels are predicted accurately (See Fig.~\ref{fig:qual_ltaa_bf} and Fig.~\ref{fig:qual_ltaa_salads}). We believe this could be because the visual information pertaining to the information is limited or negligible towards the beginning and end of the action instance.

Most classification errors result from the model getting confused among semantically similar classes. Some such cases from our examples are \textit{`take ladle'} and \textit{`pick-up ladle'} in Fig.~\ref{fig:qual_ltaa_ek}(b)); \textit{`close sandwich'} and \textit{`close hamburger'} in Fig.~\ref{fig:qual_ltaa_ek}(d)); \textit{`put seasoning'} and \textit{`pour seasoning'} in Fig.~\ref{fig:qual_ltaa_egtea}(a)).

Moreover, our model sometimes misses rare actions during predictions such as \textit{`pour oil'} in Fig~\ref{fig:qual_ltaa_ek}(a) and \textit{`close fridge'} in Fig.~\ref{fig:qual_ltaa_egtea}(b).

Additionally, we also observe that having seen certain objects in the observed video, the model predicts objects that are likely to co-occur with the seen objects. See the scenario in Fig.~\ref{fig:qual_ltaa_salads}(d). The model doesn't predict \textit{`cut\_cheese'} and \textit{`place\_cheese\_into\_bowl'} after the action \textit{`place\_cucumber\_into\_bowl'} and instead predicts \textit{cut\_tomato} and \textit{`place\_tomato\_into\_bowl'}. While the prediction is not correct for this specific activity, it is still a reasonable sequence of actions as there are several other salad recipe videos in the dataset that only use \textit{cucumber} and \textit{tomato}. In another scenario in Fig.~\ref{fig:qual_ltaa_egtea}(b), having seen \textit{`pasta'} in the observed video, the model anticipates action classes with \textit{`cheese'} noun. While \textit{`cheese'} does not appear in this particular video, it is a reasonable prediction since the nouns \textit{`pasta'} and \textit{`cheese'} often appear together in activity videos in this dataset.

\subsection{Additional Discussion}
\label{sec:discussion}
In this work, we demonstrate the effectiveness of our model on minutes-long activity videos. Handling longer videos with durations in hours or days (common in surveillance or monitoring scenarios) would be interesting future work. Furthermore, our approach assumes that the videos have an overall context provided by the ongoing long-term activity. We show that modeling interactions among segments (and, in turn, segment-level representation) is an effective technique for such activity videos as the video segments are indeed related. However, such approaches cannot tackle videos that are just a montage of several unrelated content like videos containing clips from different movies. Our approach focuses on activity videos where contextual information is present and relevant for action anticipation.

\end{document}